\documentclass[preprint,12pt]{elsarticle}
\usepackage{graphicx}
\usepackage{subcaption}
\usepackage{multirow}
\usepackage{enumitem}
\usepackage{marvosym}
\usepackage{amsmath}
\usepackage{bm}
\usepackage{xcolor}
\usepackage[linesnumbered,ruled,vlined]{algorithm2e}

\usepackage{amssymb}

\journal{Journal of Biomedical Informatics}

\begin{document}

\begin{frontmatter}

\title{ACDNet: Attention-guided Collaborative Decision Network for Effective Medication Recommendation}

\author[label1]{Jiacong Mi}
\author[label1]{Yi Zu}
\author[label1]{Zhuoyuan Wang}
\author[label1]{Jieyue He\corref{cor1}}
\ead{jieyuehe@seu.edu.cn}
\affiliation[label1]{organization={School of Computer Science and Engineering, Key Lab of Computer Network and Information Integration, MOE, Southeast University},
          city={Nanjing},
          postcode={210018},
          state={Jiangsu},
          country={China}}

\cortext[cor1]{Corresponding author}

\begin{abstract}
Medication recommendation using Electronic Health Records (EHR) is challenging due to complex medical data. Current approaches extract longitudinal information from patient EHR to personalize recommendations. However, existing models often lack sufficient patient representation and overlook the importance of considering the similarity between a patient's medication records and specific medicines. Therefore, an Attention-guided Collaborative Decision Network (ACDNet) for medication recommendation is proposed in this paper. Specifically, ACDNet utilizes attention mechanism and Transformer to effectively capture patient health conditions and medication records by modeling their historical visits at both global and local levels. ACDNet also employs a collaborative decision framework, utilizing the similarity between medication records and medicine representation to facilitate the recommendation process. The experimental results on two extensive medical datasets, MIMIC-III and MIMIC-IV, clearly demonstrate that ACDNet outperforms state-of-the-art models in terms of Jaccard, PR-AUC, and F1 score, reaffirming its superiority. Moreover, the ablation experiments provide solid evidence of the effectiveness of each module in ACDNet, validating their contribution to the overall performance. Furthermore, a detailed case study reinforces the effectiveness of ACDNet in medication recommendation based on EHR data, showcasing its practical value in real-world healthcare scenarios.
\end{abstract}

\begin{keyword}


Medication recommendation  \sep Data mining 
\sep Electronic Health Record \sep Attention mechanism 
\sep Transformer
\end{keyword}

\end{frontmatter}



\section{Introduction}
Medication recommendation is an important part of healthcare, helping to find the best treatment plans by comparing patient histories, medicine information, and other data with medical knowledge. It uses technologies like machine learning, natural language processing, and data mining to improve treatment, reduce bad drug reactions, and boost patient well-being. Electronic health records (EHR) \cite{menachemi2011benefits} have led to large, complex data sets with patient details, reports, and medication records. Medication records are key parts of EHR, providing vital info for doctors. As a result, using EHR data for medication recommendation is a main focus in recent medical studies.

In recent years, with the continuous development of machine learning and deep learning, medication recommendation models based on deep learning have attracted increasing attention \cite{ali2023deep}. These models can be divided into instance-based \cite{gong2021smr,zhang2017leap} and longitudinal-based \cite{choi2016retain,ren2022drug,shang2019gamenet,tan20224sdrug,wang2022ffbdnet,yang2021micron,yang2021safedrug,wu2022leveraging,li2023dgcl} categories. Instance-based medication recommendation models solely rely on patients' current health information, neglecting past health data and medication records, which limits their effectiveness. In contrast, longitudinal-based models provide a more comprehensive understanding of patients' conditions, thereby improving the effectiveness of medication recommendation. However, longitudinal-based medication recommendation models still face challenges and problems. Firstly, certain studies \cite{shang2019gamenet,yang2021safedrug,wu2022leveraging} utilize GRU or RNN to encode patients' EHR information, which makes it challenging to capture the comprehensive health information of patients from a global perspective. Secondly, some studies \cite{wang2022ffbdnet,yang2021micron,ren2022drug} disregard the importance of patients' medication records, overlooking the crucial combinations of medications that significantly impact patients' well-being.

To address these challenges, we propose the Attention-guided Collaborative Decision Network (ACDNet) model for medication recommendation. ACDNet utilizes attention mechanism and Transformer to effectively capture important information from the patient's historical visits and medication records at both the local and global levels, thereby providing more comprehensive medication recommendations. Firstly, ACDNet employs attention mechanism and Transformer to encode patients' visit information at both the local and global levels, generating representations for their health conditions and medication records. Secondly, it incorporates various domain-specific knowledge to comprehensively model medications. Finally, ACDNet utilizes collaborative decision network to recommend the most suitable medications for patients based on their medical history and the similarity between their medication records and available medicines.

In addition to the widely used MIMIC-III dataset \cite{johnson2016mimic}, we conducted experiments on the latest MIMIC-IV dataset \cite{johnson2020mimic}. Compared to MIMIC-III, MIMIC-IV offers more abundant medical data. The experimental results indicate that ACDNet model demonstrates significant improvements in accuracy and effectiveness compared to existing models in medication recommendation. Our main contributions are summarized as follows:
\begin{itemize}
	\item[$\bullet$]We employ attention mechanism and Transformer to effectively capture the health conditions and medication records of a patient by modeling his historical visits at both local and global levels.
	\item[$\bullet$] We incorporate a collaborative decision mechanism to provide personalized medication recommendation by leveraging various medical information from a patient’s history of visits, as well as diverse medicine information. 
	\item[$\bullet$] We conduct extensive experiments on both the MIMIC-III and MIMIC-IV datasets, demonstrating the effectiveness of our proposed approach in comparison to existing methods.
\end{itemize}

\section{Related Works}
In recent years, various medication recommendation methods \cite{wang2022ffbdnet,yang2021safedrug} have incorporated molecular information as external knowledge to improve the accuracy and effectiveness of medication recommendation. In this section, we discuss two research directions: medication recommendation and molecular representation.

\subsection{Medication recommendation}
Medication recommendation can be classified into instance-based and longitudinal-based categories. The former recommends medicines based on patients' current visit information, while the latter utilizes patients' historical visit information to predict future medication requirements. In instance-based medication recommendation, LEAP \cite{zhang2017leap} employs RNN and attention mechanism to model current information, introducing reinforcement learning to prevent adverse drug combinations. SMR \cite{gong2021smr} uses EHR data and medical knowledge graphs to construct a heterogeneous graph, transforming the medication recommendation problem into a link prediction problem. In longitudinal-based medication recommendation, RETAIN \cite{choi2016retain} proposes a two-level neural attention model effectively capturing relevant information in patients' visits. GAMENet \cite{shang2019gamenet} uses RNN to model visit information, incorporating drug interaction information to represent medications. MICRON \cite{yang2021micron} employs residual networks to predict changes in health conditions, focusing on analyzing the differences between two consecutive visits. SafeDrug \cite{yang2021safedrug} is the first to model drug molecules and proposes a new loss function to reduce drug interaction rates. COGNet \cite{wu2022conditional} employs Transformer to encode visit and medicine information, comparing current visit information with previous visits to make comprehensive judgments. DRMP \cite{ren2022drug} uses GRU and graph neural networks to capture relationships between visits, introducing a DDI Gating Mechanism to decrease drug interaction rates. FFBDNet \cite{wang2022ffbdnet} incorporates various external knowledge sources and employs a binary decision network for medication recommendation. Despite the progress achieved by existing medication recommendation methods, challenges and shortcomings still persist in adequately representing patient information and providing effective medication recommendations for patients.

\subsection{Molecular Representation}
Molecular representation transforms structures into numerical vectors for predicting properties and functions. Presently, graph neural network (GNN)-based methods \cite{wang2022molecular,wieder2020compact,zhou2020graph} are considered effective means for learning high-quality molecular representations. Both MPNN \cite{gilmer2017neural} and Chemprop \cite{yang2019analyzing} are GNN methods based on message-passing mechanism. The former updates atomic embedding vectors by exchanging information between atoms and bonds, while the latter uses self-supervised learning to improve the quality of molecular representations, learning atom-level representations by predicting partial charges for each atom in the molecule and averaging or summing these representations to form molecule-level representations. D-GCAN \cite{sun2022prediction} is a deep learning method based on graph convolutional attention networks, combining graph convolution and attention mechanism to extract feature vectors for atoms and bonds in molecular structures. D-GCAN uses graph convolution for molecular representation and then employs graph attention to calculate the importance of atoms and bonds within the molecule, ultimately forming the molecular representation. Current molecular representation techniques may encounter difficulties in capturing complex molecular features and interdependencies, necessitating further research to enhance the accuracy of predicting molecular properties and functions.

\section{Problem Formulation}
In medication recommendation, we utilize patient EHR data and external medicine information to provide personalized medicine suggestions for patients. Below are the definitions of EHR data, medicine information, and medication recommendation.

\subsection{Electronic health records (EHR)} In the EHR dataset, each patient is represented by sequential data ${X}^{(n)}=\left[{x}_{1}^{(n)}, {x}_{2}^{(n)}, \cdots, {x}_{T^{(n)}}^{(n)}\right]$. Each record ${x}_{t}=\left[c_{t}^{d}, c_{t}^{p}, c_{t}^{m}\right](t<T)$ contains diagnosis codes $c^d_t$, procedure codes $c^p_t$, and medication codes $c^m_t$. The current medical record ${x}_{T}=\left[c_{T}^{d}, c_{T}^{p}\right]$ consists of diagnosis codes $c^d_T$ and procedure codes $c^p_T$ but excludes medication codes. For simplicity, medical code $c_{t}^{*} \in\mathbb{N}^ {l^{*}_t}$  is utilized to represent the standardized definition of various medical code types where $\mathbb{N} \in [0,|C^{*}|]$, $C^{*}$ is the medical code set and $l^{*}_t$ is the number of medical codes in  $c_{t}^{*}$.

\subsection{EHR and DDI Graph} The EHR graph ${G}^{E}=\left\{{V}^{{E}}, {E}^{{E}}\right\}$ and DDI graph ${G}^{D}=\left\{{V}^{{D}}, {E}^{{D}}\right\}$ both have node sets $V^{E}=C^{M}$ and $V^{D}=C^{M}$, representing all medicines. $E^E$ corresponds to known medication combinations in the EHR database, while $E^D$ represents known drug-drug interactions from an external knowledge base. The adjacency matrices $A^{E}, A^{D} \in R^{\left|C^{M}\right| \times\left|C^{M}\right|}$ help clarify the construction of edge sets $E^E$ and $E^D$, respectively.
\subsection{Molecular Graph} Molecular graph for medication $m_i\in C^M$ can be represented as $G^{m_i}=\left\{V^{m_i},E^{m_i}\right\}$, where $V^{m_i}$ is the node set consisting of all molecular units of medication $m_i$, and $E^{m_i}$ is the edge set of known molecular structures of $m_i$. So the adjacency matrix for each medicine molecule can be represented as $A^{m_i}$.
\subsection{Medication Recommendation} Given the current time $T$,  medical codes  ${x}_T=\left[c_T^d,c_T^p\right]$ that only include diagnoses and procedures, as well as the  medical records ${X}=\left[{x_1},{x_2},\cdots,{x}_{T-1}\right]$ from the previous $T-1$ times and graphs $G^{E}, G^{D},\left\{G^{m_{i}}, {i} \in\left[1, \ldots,\left|C^{M}\right|\right]\right\}$, the aim is to obtain a multi-class output $ {\hat{o}}\in\{0,1\}^{\left|C^m\right|}$ of medicines while minimizing the interactions between them.

\section{Method}

\begin{figure}[h]
	\centering
	\includegraphics[width=1.0\linewidth]{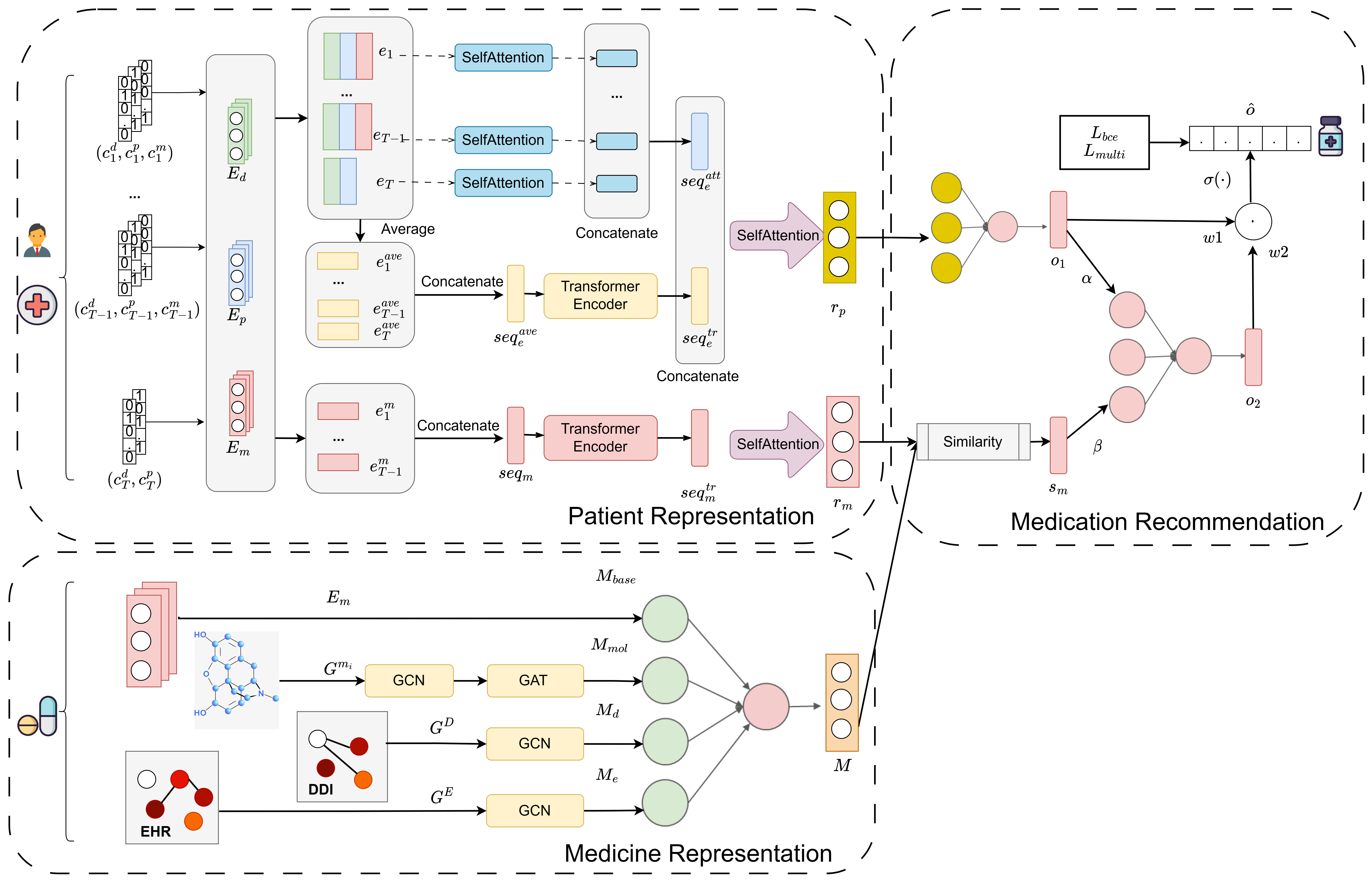}
	\caption[]{The architecture of ACDNet}
	\label{fig1}
\end{figure}

Our ACDNet model has three modules: a patient representation module, a medicine representation module, and a medication recommendation module. In the patient representation module, we utilize attention mechanism and Transformer encoder to extract essential information from historical medical records, creating the patient's representations of their past visits and medication history from both local and global perspectives. The medicine representation module encodes the medicine molecular graph and uses the EHR and DDI graphs to generate medicines' vector representation. Finally, the medication recommendation module employs a collaborative decision mechanism to determine the medication to recommend for the patient. The ACDNet model's architecture is shown in Figure ~\ref{fig1}.
\subsection{Patient representation}
By leveraging a patient's past health information and medication records, we can gather a significant amount of information related to their well-being. During the ongoing medical consultation, we further determine the patient's latest health condition. By integrating the collected data, we can conduct a thorough evaluation of the patient's health and provide customized medication recommendation. Specifically, we initially represent each visit as a vector using the EHR Embedding module. Following this, with the application of techniques such as attention mechanism and Transformer encoder, we obtain encoded representations of visit sequences and medication records sequences. Finally, by employing the self-attention mechanism, we generate a vector representation of the patient's historical health and medication records, establishing a foundation for subsequent personalized recommendation.
\subsubsection{Attention  module}
We employ the Self-Attention mechanism to effectively capture valuable information from a sequence vector of length $L$. This mechanism allows us to focus on the most relevant parts of the sequence, considering the relationships and dependencies among different elements, ultimately enhancing our ability to extract significant insights from the sequence data.

	\begin{equation}
		\text{SelfAttention}(v) = \text{softmax}\left(tanh(v \cdot W_{\alpha1})W_{\alpha2}\right)^T \cdot v
	\end{equation}
	where $v \in R^{L \times dim}$ represents the new input sequence, $W_{\alpha1} \in R^{dim \times dim}$, $W_{\alpha2} \in R^{dim \times 1}$. The output dimension of the sequence is $R^{dim}$.
	
\subsubsection{Transformer encoder module} 
During the encoding process, the Transformer encoder \cite{vaswani2017attention} can capture important information from the historical visit records and medication records. The Transformer encoder consists of $L$ Transformer encoder Layers, and for the $i^{th}$ layer, its output is:
\begin{equation}
	Z^{(i)}=\text {LayerNorm}\left(\text {MultiHeadAttention}\left({seq}^{(i)}, {seq}^{(i)}, {seq}^{(i)}\right)+{seq}^{(i)}\right) \\
\end{equation}
\begin{equation}
	F^{(i)}=\text{LayerNorm}\left(\text {FeedForward}\left(Z^{(i)}\right)+Z^{(i)}\right)
\end{equation}

	Describing MultiHeadAttention, we have the following expressions:
	\begin{equation}
		\text{MultiHeadAttention}(Q,K,V)=\text{Concat}(\text{head}_1,\text{head}_2,\ldots,\text{head}_h)W^O
	\end{equation}
	\begin{equation}
		\text{Attention}(Q,K,V)=\text{softmax}\left(\frac{QK^T}{\sqrt{d_k}}\right)V
	\end{equation}
Here, $\mathrm{head}_i=\mathrm{Attention}(Q_i, K_i, V_i)$ denotes the computation for each attention head, where $\sqrt{d_k}$ serves as the scaling factor, and $W^O$ represents the output weight matrix utilized for the linear transformation and aggregation of results from various attention heads to yield the ultimate output.

The final output of the Transformer encoder is:
\begin{equation}
	\text {TransformerEncoder}({seq})=\text {LayerNorm}\left({seq}+\text{sum}\left(F^{(1)}, \ldots, F^{(L)}\right)\right)
\end{equation}

	where sum is the pooling operation that involves adding up the results.

\subsubsection{Local representation Module}
Based on the previous context, the vector $x_t = [c_t^d, c_t^p, c_t^m]$ represents the representation of a visit. To refine the processing of these vectors, we have devised three embedding operations, $E_d: {\mathbb{N}^{l^{d}_t}} \rightarrow R^{{l^{d}_t} \times dim}$, $E_p : \mathbb{N}^{l^p_t} \rightarrow {R}^{l^p_t \times dim}$, and $E_m : \mathbb{N}^{l^m_t} \rightarrow {R}^{l^m_t \times dim}$, facilitating their mapping into the embedding space. Explicitly, upon querying each vector within the embedding operations, we obtain the respective embedded vectors, culminating in the subsequent result:
\begin{equation}
\begin{aligned}
	e_{t}^{d}&=E_{d} (c_{t}^{d}) \\
	e_{t}^{p}&=E_{p} (c_{t}^{p}) \\
	e_{t}^{m}&=E_{m} (c_{t}^{m})
\end{aligned}
\label{emb}
\end{equation}

To obtain the sequential representation of the visit $x_t$, we concatenate the following three vectors:
\begin{equation}
	e_{t} = \left[e_{t}^{d} \, \|\, e_{t}^{p} \, \|\, e_{t}^{m}\right]
\end{equation}
where $e_{t}\in R^{l\times dim}$, $l$ is the sum of ${l^d_t},{l^p_t}$ and ${l^m_t}$. Consequently, by executing an averaging operation on this vector, we ascertain the current visit's average representation:
\begin{equation}
	e_{t}^{ave}={average}\left(e_{t}\right)
\end{equation}
where $e_{t}^{ave}\in R^{dim}$ and $average$ denotes the vector's averaging operation.

Our goal is to enhance the performance of the model by capturing the interaction information between medical codes. To achieve this, we employ self-attention mechanism to calculate the interdependencies among the medical code features within the current input sequence, in order to capture the important information from each patient visit. Thus, by applying the self-attention mechanism, we can model the informatiovn of a patient's visit at time $t$ and obtain a local representation of the patient's visit history:
\begin{equation}
	v_{t}^{a t t}=\text{SelfAttention}\left(e_{t}\right)
\end{equation}
where $v_{t}^{a t t} \in R^{dim}$. Subsequently, we concatenate vectors to acquire the attention representation sequence for the patient's $T$ visits:
\begin{equation}
	{seq}_{e}^{att}=\left[ v_{1}^{att} || v_{2}^{att} || \ldots || v_{T}^{a t t}\right ]
\end{equation}
where ${seq}_{e}^{att} \in R^{T \times dim}$.
\subsubsection{Global representation module} 

In COGNet \cite{wu2022conditional}, the model employs the Transformer to encode medical records for individual visits, with a specific focus on capturing the relationships between various medical codes within each visit. In contrast, ACDNet treats each visit as a vector, arranging all visits into a sequence processed through a Transformer encoder to emphasize the relationships between distinct visits.

We handle the complete set of patient visit and medication records separately, enabling the capture of essential information at a global level throughout the patient's entire visit history. The patient's health status and medication records after $T$ visits can be represented as:
\begin{equation}
	seq_{e}^{ave}=\left [e_{1}^{a v e} || e_{2}^{a v e} || \ldots || e_{T}^{a v e}\right ]
\end{equation}
\begin{equation}
	{seq}_{m}=\left [e_{1}^{m} || e_{2}^{m} || \ldots || e_{T-1}^{m}\right ]
\end{equation}
where ${seq}_{e}^{ave} \in R^{T \times dim}$ and ${seq}_{m} \in R^{(T-1) \times dim}$. Then, we input them into the Transformer encoder separately to obtain the encoded vectors:
\begin{equation}
	{seq}_{e}^{t r}=\text {TransformerEncoder}\left({seq}_{e}^{a v e}\right)
\end{equation}
\begin{equation}
	{seq}_{m}^{t r}=\text{TransformerEncoder}\left({seq}_{m}\right)
\end{equation}
where ${seq}_{e}^{tr} \in R^{T \times dim}$ and ${seq}_{m}^{tr} \in R^{(T-1) \times dim}$.
\subsubsection{Represent module}
After processing with the attention mechanism and the Transformer encoder, we can obtain the representations of the patient's health status and medication records after $T$ medical visits. For the patient's health information representation, we concatenate the two patient health status sequences and perform a self-attention operation:
\begin{equation}
	r_{p}=\text{SelfAttention}\left({seq}_{e}^{a t t} || s e q_{e}^{t r}\right) 
\end{equation}
where $r_{p} \in R^{dim}$. Similarly, we perform a self-attention operation on the medication records sequence:
\begin{equation}
	r_{m}=\text{SelfAttention}\left({seq}_{m}^{tr}\right) 
\end{equation}
where $r_{m} \in R^{dim}$.
\subsection{Medicine Representation}
To improve the effectiveness of medication recommendation and better express medicine information, we employ various methods to model medicines, including EHR graph, DDI graph, and molecular graphs. First, we use a basic embedding table to represent all medicines. Then, we introduce graph neural networks to encode the graph information. Finally, we aggregate all representation information to obtain the final medicine representation. As shown in Equation (\ref{emb}), we use the learned Embedding table $E_m$ as the base representation for all medicines:
\begin{equation}
	M_{base}=E_{m} (C^{M})
\end{equation}
where $M_{base} \in R^{|C^m|\times dim}$. 

As mentioned earlier, we obtained the EHR Graph and DDI Graph with their adjacency matrices represented as $A^E$ and $A^D$, respectively. Inspired by the use of GCN \cite{kipf2016semi} to encode graph information in GAMENet \cite{shang2019gamenet}, we apply GCN separately to the EHR graph and DDI graph, given the graph adjacency matrix $A^*$:
\begin{equation}
	\tilde{A}^{*}=\widetilde{D}^{-\frac{1}{2}}\left(A^{*}+I\right) \widetilde{D}^{-\frac{1}{2}}
\label{astar}
\end{equation}
where $\widetilde{D}$ is a diagonal matrix of $A^*$ and $I$ are identity matrices. Then, we can obtain the graph representation:
\begin{equation}
	M^{*}=\tilde{A}^{*} \sigma\left(\tilde{A}^{*} W_{g 1}^{*}\right) W_{g 2}^{*}
\label{mstar}
\end{equation}
where $\sigma$ is a nonlinear activation function, $ W_{g1}^{*}\in R^{|V^{*}|\times dim}$, $W_{g2}^*\in R^{dim\times dim}$ are learnable parameters. Based on the above formula, we obtain the EHR Graph representation $M_e\in R^{|C^m|\times dim}$ and DDI Graph representation $M_d\in R^{|C^m|\times dim}$.

Medicine molecules can be represented in the form of molecular graphs, where nodes represent atoms and edges represent chemical bonds between atoms. Inspired by D-GCAN \cite{sun2022prediction}, we adopt GCN \cite{kipf2016semi} and GAT \cite{velivckovic2017graph} to represent molecules. For each molecule $G^{m^i}$ with adjacency matrix $A^{m^i}$, we first obtain $M_g^{m^i}\in R^{|V^{m^i}|\times dim}$ through equations (\ref{astar}) and (\ref{mstar}). Then, using GAT and Readout fuction, we get:
\begin{equation}
	M^{m^{i}}= \operatorname{READOUT}\left(GAT\left(M_{g}^{m^{i}}\right)\right)
\end{equation}
where $M^{m^i}\in R^{dim}$ is the vector representation of the $i^{th}$ medicine molecule. Finally, we obtain the representations of all molecules:
\begin{equation}
	M_{m o l}=\left[M^{m^{1}} || M^{m^{2}} || \ldots || M^{\left|C^{M}\right|}\right]
\end{equation}
where $M_{mol}\in R^{|C^M|\times dim}$ represents the molecular graph information of all drugs. Ultimately, we obtain the final medicine representation:
\begin{equation}
	M=\text{NN}_{\text {med }}\left(M_{\text {base }} || M_{e} || M_{d} || M_{\text {mol}}\right)
\end{equation}
where $\text{NN} _{\text {med }}(\cdot): R^{4 d i m} \rightarrow R^{d i m}$ is a fully connected feedforward neural network, and $M\in R^{|C^M|\times dim}$ incorporates various medicine information, which is beneficial for similarity calculations with the patient's medication information.
\subsection{Medication Recommendation}

Drawing inspiration from FFBDNet\cite{wang2022ffbdnet}, our medication recommendation system incorporates two modules that work collaboratively to provide medication suggestions for patients. The direct module primarily focuses on the patient's representation to recommend suitable medicines. On the other hand, in contrast to FFBDNet, the indirect module takes into account the similarity between the patient's medication records and medicine representation. By integrating the patient's medication history, which contains valuable information about relevant medicines, we aim to simulate the decision-making process of a prescribing doctor. This integration enables us to generate comprehensive personalized medication recommendation based on the patient's specific medication history.

In the direct recommendation module, we use the patient representation and a feedforward neural network to recommend suitable medicines directly:
\begin{equation}
	o_{1}=\text{NN}_{o_{1}}\left(r_{p}\right)
\end{equation}
where $o_1\in R^{|C^m|}$ is the result of the direct recommendation, and $N N_{o_{1}}(\cdot): R^{d i m} \rightarrow R^{\left|C^{m}\right|}$ is a learnable feedforward neural network.

The indirect recommendation module integrates the patient's medication records in EHR with the representations of medicines and calculates the similarity between them. This similarity calculation mimics the prescribing behavior of healthcare professionals, selecting the most important combination of medications from all the patient's medication records.  
\begin{equation}
	s_{ m}=\text { Cosine Similarity }(r_{m}, M)=\frac{r_{m} \cdot M^{T}}{\|r_{m}\| \cdot\|M\|}
\end{equation}
where $s_{m}\in R^{|C^m|}$ is the similarity between medication records and medicines. 

Then, we concatenate the direct recommendation $o_1$ and similarity results and input them into a neural network to obtain the indirect recommendation result:
\begin{equation}
	o_{2}=\text{NN}_{o_{2}}\left(\alpha o_{1} || \beta s_{m} \right)
\end{equation}
where $o_2\in R^{|C^m|}$ is the output result of indirect recommendation based on similarity, $\text{NN}_{o_{2}}(\cdot): R^{2 \times\left|C^{m}\right|} \rightarrow R^{\left|C^{m}\right|}$ is a fully connected feedforward neural network, and $\alpha$, $\beta\in R^1$ are trainable parameters.

Finally, we adjust the two modules to obtain the ultimate recommendation result:
\begin{equation}
	\hat{o}=\sigma \left(w_{1} \odot o_{1}+w_{2} \odot o_{2}\right)
\end{equation}
where $\sigma$ is a nonlinear activation function, $w_1, w_2\in R^{|C^m|}$ are learnable parameters to adjust the two recommendation modules, and $\hat{o}\in R^{|C^m|}$ is the output medication recommendation result.

\subsection{Training and Inference}

To suggest medicine combinations, we treat this task as a multi-label prediction problem. During training, we use two common multi-label loss functions: binary cross-entropy loss ($L_{bce}$) and multi-label margin loss ($L_{multi}$). Binary cross-entropy loss handles multi-label binary classification problems by calculating binary cross-entropy loss for each label and summing them up. Multi-label margin loss accounts for label correlations, producing smoother predictions. 
\begin{equation}
	L_{b c e}=- {\textstyle \sum_{i}^{\left|C^{m}\right|}} y_{i} \log \left(\hat{o}_{i}\right)+\left(1-y_{i}\right) \log \left(1-\hat{o}_{i}\right)
\end{equation}
\begin{equation}
	L_{\text {multi }}=\sum_{i, j: y_{i}=1, y_{j}=0} \frac{\max \left(0,1-\left(\hat{o}_{\mathrm{i}}-\hat{o}_{\mathrm{j}}\right)\right)}{|y|}
\end{equation}
where $\mathbf{y} \in\{0,1\}^{\left|C^{m}\right|}$ is the true label set of recommended medicines. Therefore, the final Loss is represented as:
\begin{equation}
	{L}=\lambda L_{b c e}+(1-\lambda) L_{\text {multi}}
\end{equation}
where $\lambda \in R$ is a predefined parameter.

In ACDNet training, we combine these loss functions for better prediction performance, as detailed in \textbf{Algorithm 1}.

\begin{algorithm}[h]
	\caption{One training epoch of ACDNet}\label{algorithm}
	\KwIn{Training set $X^{(n)}=\left[x_{1}^{(n)}, x_{2}^{(n)}, \cdots, x_{T^{(n)}}^{(n)}\right]$, EHR graph $A^E$, DDI graph $A^D$, molecular graph $\left\{A^{m_{i}}, i \in\left[1,\ldots,\left|C^{m}\right|\right]\right\}$}
	Generate the medicine representation $M$ according to Equations (18)-(23)\;
	\For{patient $i=1$ to $N$}{
		Read the patient $i$’s history $x_1$, $x_2$, ... , $x_T$\;		 
		\For{history visit $t=1$ to $T$}{
			Read the historical diagnoses, procedures, medicines of the patient at the $t^{th}$ visit $\left[c_{1}^{d}, c_{2}^{d}, \ldots c_{t}^{d}\right],\left[c_{1}^{p}, c_{2}^{p}, \ldots c_{t}^{p}\right],\left[c_{1}^{m}, c_{2}^{m}, \ldots c_{t-1}^{m}\right]$\;
			Generate embeddings $e_t$ and $e_t^{ave}$ by Equations (7)-(9)\;
			Generate visit $v_t^{att}$ using self-attention by Equation (10)\;
			Generate sequences $seq_e^{att}, seq_e^{tr}, seq_m^{tr}$ by Equations (11)-(15)\;
			Generate patient representation $r_p$ and medication records representation $r_m$ by
			Equations (16)-(17)\;
			Generate similarity $s_{m}$ by Equations (25)\;
			Generate output $\hat{o}$ by Equations (24), (26)-(27)\;
		}
		Generate and accumulate $L_{bce}$, $L_{multi}$ in Equations (28)-(29)\;
	}
	Optimize the combined loss $L$ in Equation (30)
\end{algorithm}

\section{Experiment}
To validate the effectiveness of ACDNet, we conducted experiments on two publicly available datasets, which can be divided into the following three sections:
\begin{itemize}
	\item[$\bullet$] In comparative experiments, we assessed the performance of ACDNet and baselines on both datasets.
	\item[$\bullet$] In the parameter experiments, we explore different parameter combinations in the parameter experiments to find the optimal model configuration for ACDNet.
	\item[$\bullet$] In ablation experiments, we separately verified the effectiveness of each module in ACDNet.
	\item[$\bullet$] Additionally, we performed case validation, demonstrating the practical application of ACDNet in real-world scenarios.
\end{itemize}
\subsection{Dateset}
We employed MIMIC-III \cite{johnson2016mimic} and MIMIC-IV \cite{johnson2020mimic} as sources for our experimental data. For the MIMIC-IV dataset, diagnoses and procedures are based on ICD-9 and ICD-10 codes. To process this information, we initially converted ICD-10 coded data to ICD-9 coded data \cite{national2016icd}. In accordance with the data processing methods utilized in GAMENet \cite{shang2019gamenet} and SafeDrug \cite{yang2021safedrug}, we processed the datasets, resulting in the data presented in Table \ref{tab1}.

\begin{table}[h]
	\caption{Statistics of the datasets}\label{tab1}
	\renewcommand{\arraystretch}{1.2}
	\scriptsize
	\centering
	\begin{tabular}{lcc}
		\hline
		& MIMIC-III & MIMIC-IV \\ \hline
		\# patients              & 6350      & 64462    \\
		\# clinical events       & 15032     & 171295   \\
		\# diagnoses             & 1958      & 2000     \\
		\# procedures            & 1430      & 1500     \\
		\# medicines             & 131       & 131      \\ \hline
		Avg/max \# of visits     & 2.37/29   & 2.66/66  \\
		Avg/max \# of diagnoses  & 10.51/128 & 8.54/228 \\
		Avg/max \# of procedures & 3.84/50   & 2.22/72  \\
		Avg/max \# of medicines  & 11.44/65  & 6.54/72  \\ \hline
		total \# of DDI pairs    & 448       & 448      \\ \hline
	\end{tabular}
\end{table}

\subsection{Baselines and Metrics}
We compare our model with the baselines below.
\begin{itemize}
	\item[$\bullet$] LR is a instance-based classifier with L2 regularization.
	\item[$\bullet$]  ECC \cite{read2011classifier} is an ensemble model for multi-label classification with correlated labels.
	\item[$\bullet$] RETAIN \cite{choi2016retain} is an interpretable healthcare model using reverse time attention.
	\item[$\bullet$] LEAP \cite{zhang2017leap} captures label dependencies and prevents adverse drug interactions with reinforcement learning.
	\item[$\bullet$] GAMENet \cite{shang2019gamenet} recommends safe medication combinations using graph-augmented memory networks.
	\item[$\bullet$] SafeDrug \cite{yang2021safedrug} is a DDI-controllable drug recommendation model utilizing molecular structures.
	\item[$\bullet$] MICRON \cite{yang2021micron} predicts medication changes based on patient health using recurrent residual networks.
	\item[$\bullet$] COGNet \cite{wu2022conditional} generates medicine sets for patients based on diagnoses with a copy-or-predict mechanism.
	\item[$\bullet$] DRMP \cite{ren2022drug} is a drug recommendation model using message propagation and DDI gating mechanism.
	\item[$\bullet$] FFBDNet \cite{wang2022ffbdnet} is a feature fusion and bipartite decision network for medication combination recommendation.
\end{itemize}

We evaluate the prediction performance using several metrics, including DDI rate, Jaccard similarity score (Jaccard) \cite{niwattanakul2013using}, Average F1 score (F1), PR-AUC \cite{davis2006relationship}, precision@k and nDCG@k, to assess the safety, accuracy, and effectiveness of our model.
\subsection{Implementation details}
In line with previous research efforts \cite{shang2019gamenet,yang2021safedrug}, we divided the dataset into training, validation, and test sets at a 4:1:1 ratio. Our model is implemented using Python 3.9.15 and PyTorch 1.12.1 \cite{paszke2017automatic}, with training and testing conducted on Intel Xeon CPU and NVIDIA 3080Ti GPU. After conducting parameter experiments, optimal hyperparameters were selected within the model, with the embedding dimension size set at 64 and $\lambda$ set at 0.97. For the Transformer encoder, we selected a configuration with 8 attention heads and 6 layers. Due to varying dataset sizes, a learning rate of 0.0015 was established for the MIMIC-III dataset, and 0.0002 for the MIMIC-IV dataset. We utilized the Adam optimizer to refine the model. For all baseline models, we used the optimization parameters mentioned in the literature for training and testing on the MIMIC-III dataset. Similarly, for the MIMIC-IV dataset, we re-optimized the parameters for all baseline models to ensure their optimal performance. According to SafeDrug \cite{yang2021safedrug}, in the evaluation process, instead of conducting cross-validation, we apply bootstrapping sampling. Specifically, we randomly sample 80\% of the data points from the test set for one round of evaluation. We perform 10 rounds of sampling and evaluation, and report the mean and standard deviation values.
\subsection{Performance Comparison}
Table 2 presents the performance of various models on the MIMIC-III dataset. ACDNet outperforms the baseline models in terms of Jaccard, PR-AUC, and F1 metrics, demonstrating its effectiveness in capturing both local and global patient information and facilitating collaborative decision-making. The results suggest that longitudinal-based models are more accurate than instance-based approaches, as they can better capture patients' historical health information, leading to improved patient representations. In terms of reducing the drug-drug interaction (DDI) rate, SafeDrug performs well due to its utilization of DDI loss to minimize the DDI rate. However, it is important to note that the actual DDI rate is 0.08379, and ACDNet's results closely resemble the real dataset, indicating its reliability and alignment with the genuine data.
\begin{table}[h]
	\caption{Performance comparison on MIMIC-III}\label{tab2}
	\centering
	\renewcommand{\arraystretch}{1.2}
	\setlength{\tabcolsep}{2pt}
	\scriptsize
	\begin{tabular}{lccccc}
		\hline
		Models  & Jaccard       & PR-AUC        & F1            & DDI           & Avg\_Med    \\ \hline
		LR       & 0.4865 ± 0.0021 & 0.7509 ± 0.0018 & 0.6434 ± 0.0019  & 0.0829 ± 0.0009 & 16.1773 ± 0.0942  \\
		ECC      & 0.4996 ± 0.0049 & 0.6844 ± 0.0038 & 0.6569 ± 0.0044 & 0.0846 ± 0.0018 & 18.0722 ± 0.1914  \\
		RETAIN   & 0.4887 ± 0.0028 & 0.7556 ± 0.0033  & 0.6481 ± 0.0027 & 0.0835 ± 0.0020 & 20.4051 ± 0.2832 \\
		LEAP     & 0.4521 ± 0.0024 & 0.6549 ± 0.0033  & 0.6138 ± 0.0026 & 0.0731 ± 0.0008  & 18.7138 ± 0.0666 \\
		GAMENet  & 0.5067 ± 0.0025 & 0.7631 ± 0.0030 & 0.6626 ± 0.0025 & 0.0864 ± 0.0006 & 27.2145 ± 0.1141 \\
		SafeDrug & 0.5213 ± 0.0030 & 0.7647 ± 0.0025 & 0.6768 ± 0.0027 & \textbf{0.0589 ± 0.0005} & 19.9178 ± 0.1604 \\
		MICRON   & 0.5100 ± 0.0033 & 0.7687 ± 0.0026 & 0.6654 ± 0.0031 & 0.0641 ± 0.0007 & 17.9267 ± 0.2172 \\
		COGNet   & 0.5336 ± 0.0011 & 0.7739 ± 0.0009 & 0.6869 ± 0.0010 & 0.0852 ± 0.0005 & 28.0900 ± 0.0950 \\
		DRMP     & 0.5312 ± 0.0015 & 0.7757 ± 0.0016 & 0.6854 ± 0.0011 & 0.0865 ± 0.0006 & 22.7300 ± 0.2300 \\
		FFBDNet  & 0.5292 ± 0.0020 & 0.7777 ± 0.0010 & 0.6833 ± 0.0017 & 0.0717 ± 0.0016 & 19.6900 ± 0.3000 \\
		ACDNet   & \textbf{0.5433 ± 0.0027} & \textbf{0.7904 ± 0.0021} & \textbf{0.6957 ± 0.0021} & 0.0859 ± 0.0010 & 20.4900 ± 0.1197 \\ \hline
	\end{tabular}
\end{table}

Table 3 presents the performance of various models on the MIMIC-IV dataset. Similar to the MIMIC-III dataset, longitudinal-based models prove more accurate than instance-based approaches on the MIMIC-IV dataset. Although the LEAP model performs well in DDI, its accuracy is inferior to other models. To enhance SafeDrug's accuracy, we set its DDI threshold to 0.1. The results reveal that despite SafeDrug's lower DDI value, its accuracy remains subpar compared to other models. The MICRON model primarily focuses on differences between adjacent visits and performs poorly on the more complex MIMIC-IV dataset. COGNet, DRMP, and FFBDNet exhibit satisfactory performance on the MIMIC-IV dataset, yet still fall short of ACDNet. ACDNet excels in Jaccard, PR-AUC, and F1 metrics, signifying its superior capability in obtaining patient health information and medication representations compared to other models.
\begin{table}[h]
	\caption{Performance comparison on MIMIC-IV}\label{tab3}
	\centering
	\renewcommand{\arraystretch}{1.2}
	\setlength{\tabcolsep}{2pt}
	\scriptsize
	\begin{tabular}{lccccc}
		\hline
		Models  & Jaccard       & PR-AUC        & F1            & DDI           & Avg\_Med    \\ \hline
		LR       & 0.4510 ± 0.0013 & 0.7290 ± 0.0014 & 0.6007 ± 0.0013 & 0.0762 ± 0.0004& 8.9866 ± 0.0374  \\
		ECC      & 0.4233 ± 0.0010 & 0.7284 ± 0.0012 & 0.5680 ± 0.0019 & 0.0771 ± 0.0003& 8.1070 ± 0.0235 \\
		RETAIN   & 0.4239 ± 0.0017 & 0.6798 ± 0.0018 & 0.5791 ± 0.0017 & 0.0939 ± 0.0015& 10.8602 ± 0.0736 \\
		LEAP     & 0.4287 ± 0.0012 & 0.5506 ± 0.0015 & 0.5820 ± 0.0012 & \textbf{0.0592 ± 0.0004} & 11.5198 ± 0.0459\\
		GAMENet  & 0.4507 ± 0.0013 & 0.7174 ± 0.0012 & 0.6043 ± 0.0014 & 0.0890 ± 0.0003 & 18.4426 ± 0.0474 \\
		SafeDrug & 0.4651 ± 0.0016 & 0.7118 ± 0.0016 & 0.6117 ± 0.0014 & 0.0740 ± 0.0004 & 14.4705 ± 0.0575\\
		MICRON   & 0.4554 ± 0.0026 & 0.6842 ± 0.0025 & 0.6088 ± 0.0032 & 0.0637 ± 0.0016 & 15.6963 ± 0.2875\\
		COGNet   & 0.4884 ± 0.0009 & 0.7087 ± 0.0008 & 0.6367 ± 0.0009 & 0.0894 ± 0.0003 & 19.7235 ± 0.0242 \\
		DRMP     & 0.4913 ± 0.0015 & 0.7338 ± 0.0021 & 0.6435 ± 0.0018 & 0.0872 ± 0.0013 & 15.0512 ± 0.0365\\
		FFBDNet  & 0.4970 ± 0.0010 & 0.7435 ± 0.0012 & 0.6454 ± 0.0009 & 0.0838 ± 0.0004 & 11.2906 ± 0.0658 \\
		ACDNet   & \textbf{0.5077 ± 0.0015} & \textbf{0.7501 ± 0.0017} & \textbf{0.6564 ± 0.0013} & 0.0849 ± 0.0005& 12.7024 ± 0.0005\\ \hline
	\end{tabular}
\end{table}

Table 4 and Table 5 display the precision and nDCG results on the MIMIC-III and MIMIC-IV datasets. From the results, it is evident that ADCNet performs the best in most cases, confirming its effectiveness in recommendations. LEAP does not perform well, possibly because its reinforcement learning process for sequential decisions may not effectively meet the requirements of recommendations. Similarly, the information capture between adjacent visits in MICRON does not yield strong results. In contrast, DRMP, which uses GNN and GRU to model patient information, has shown promising results.
\begin{table}[h]
	\caption{Recommendation performance comparison on MIMIC-III}\label{tabre3}
	\centering
	\renewcommand{\arraystretch}{1.2}
	\setlength{\tabcolsep}{2pt}
	\scriptsize
	\begin{tabular}{lcccc}
		\hline
		Model    & precision@5              & precision@10             & nDCG@5                   & nDCG@10                  \\ \hline
		LR       & 0.8852 ± 0.0043          & 0.8075 ± 0.0035          & 0.8982 ± 0.0045          & 0.8404 ± 0.0038          \\
		ECC      & 0.8851 ± 0.0033          & 0.8111 ± 0.0024          & 0.8993 ± 0.0038          & 0.8433 ± 0.0026          \\
		RETAIN   & 0.8913 ± 0.0046          & 0.8126 ± 0.0033          & 0.9016 ± 0.0040          & 0.8436 ± 0.0030          \\
		LEAP     & 0.7211 ± 0.0042          & 0.6772 ± 0.0040          & 0.7411 ± 0.0041          & 0.7033 ± 0.0039          \\
		GAMENet  & 0.8953 ± 0.0054          & 0.8224 ± 0.0051          & 0.9079 ± 0.0049          & 0.8534 ± 0.0046          \\
		SafeDrug & 0.8964 ± 0.0040          & 0.8200 ± 0.0023          & 0.9088 ± 0.0037          & 0.8518 ± 0.0022          \\
		MICRON   & 0.8655 ± 0.0061          & 0.8131 ± 0.0040          & 0.8740 ± 0.0054          & 0.8354 ± 0.0039          \\
		COGNet   & 0.8913 ± 0.0037          & 0.8114 ± 0.0030          & 0.9045 ± 0.0027          & 0.8450 ± 0.0024          \\
		DRMP     & 0.9003 ± 0.0032          & 0.8290 ± 0.0028          & 0.9097 ± 0.0024          & 0.8575 ± 0.0021          \\
		FFBDNet  & 0.9029 ± 0.0042          & 0.8349 ± 0.0041          & 0.9136 ± 0.0043          & 0.8633 ± 0.0040          \\
		ACDNet   & \textbf{0.9061 ± 0.0034} & \textbf{0.8400 ± 0.0026} & \textbf{0.9137 ± 0.0026} & \textbf{0.8660 ± 0.0021} \\ \hline
	\end{tabular}
\end{table}
\begin{table}[h]
	\caption{Recommendation performance comparison on  MIMIC-IV}\label{tabre4}
	\centering
	\renewcommand{\arraystretch}{1.2}
	\setlength{\tabcolsep}{2pt}
	\scriptsize
	\begin{tabular}{lcccc}
		\hline
		Model    & precision@5              & precision@10             & nDCG@5                   & nDCG@10                  \\ \hline
		LR       & 0.7910 ± 0.0019          & 0.6581 ± 0.0016          & 0.8156 ± 0.0019          & 0.7154 ± 0.0016          \\
		ECC      & 0.7858 ± 0.0017          & 0.6571 ± 0.0013          & 0.8105 ± 0.0014          & 0.7132 ± 0.0012          \\
		RETAIN   & 0.7374 ± 0.0036          & 0.6109 ± 0.0029          & 0.7594 ± 0.0035          & 0.6645 ± 0.0029          \\
		LEAP     & 0.5571 ± 0.0023          & 0.5139 ± 0.0023          & 0.5567 ± 0.0027          & 0.5275 ± 0.0024          \\
		GAMENet  & 0.7722 ± 0.0014          & 0.6398 ± 0.0011          & 0.7944 ± 0.0016          & 0.6954 ± 0.0011          \\
		SafeDrug & 0.7740 ± 0.0018          & 0.6379 ± 0.0019          & 0.7974 ± 0.0019          & 0.6955 ± 0.0019          \\
		MICRON   & 0.7583 ± 0.0022          & 0.6236 ± 0.0021          & 0.7801 ± 0.0023          & 0.6796 ± 0.0022          \\
		COGNet   & 0.7609 ± 0.0012          & 0.6359 ± 0.0014          & 0.7870 ± 0.0011          & 0.6918 ± 0.0011          \\
		DRMP     & 0.7938 ± 0.0016          & \textbf{0.6692 ± 0.0013} & 0.8150 ± 0.0017          & 0.7218 ± 0.0014          \\
		FFBDNet  & 0.7931 ± 0.0011          & 0.6546 ± 0.0017          & 0.8157 ± 0.0010          & 0.7125 ± 0.0014          \\
		ACDNet   & \textbf{0.8005 ± 0.0020} & 0.6647 ± 0.0020          & \textbf{0.8235 ± 0.0019} & \textbf{0.7221 ± 0.0019} \\ \hline
	\end{tabular}
\end{table}
\subsection{Parameter influence}
In this section, we will evaluate the performance of ACDNet on the MIMIC-III dataset using various parameter settings. Specifically, we will explore different combinations of $\lambda$ values and embedding dimensions to identify the optimal configuration that yields the best results.

\begin{figure}[h]
	\centering
	\begin{subfigure}[b]{0.3\linewidth}
		\centering
		\includegraphics[width=\linewidth]{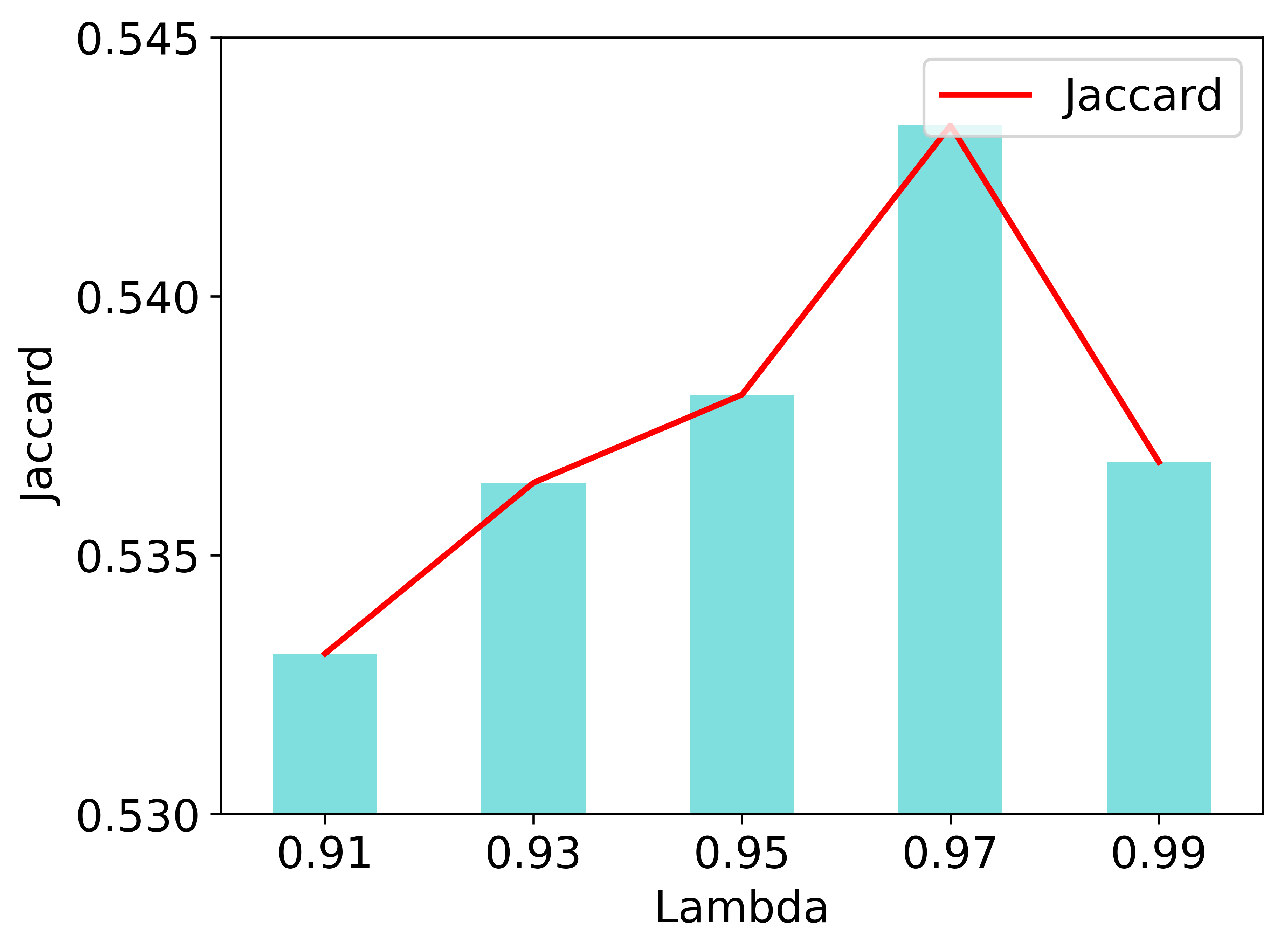}
	\end{subfigure}
	\hfill
	\begin{subfigure}[b]{0.3\linewidth}
		\centering
		\includegraphics[width=\linewidth]{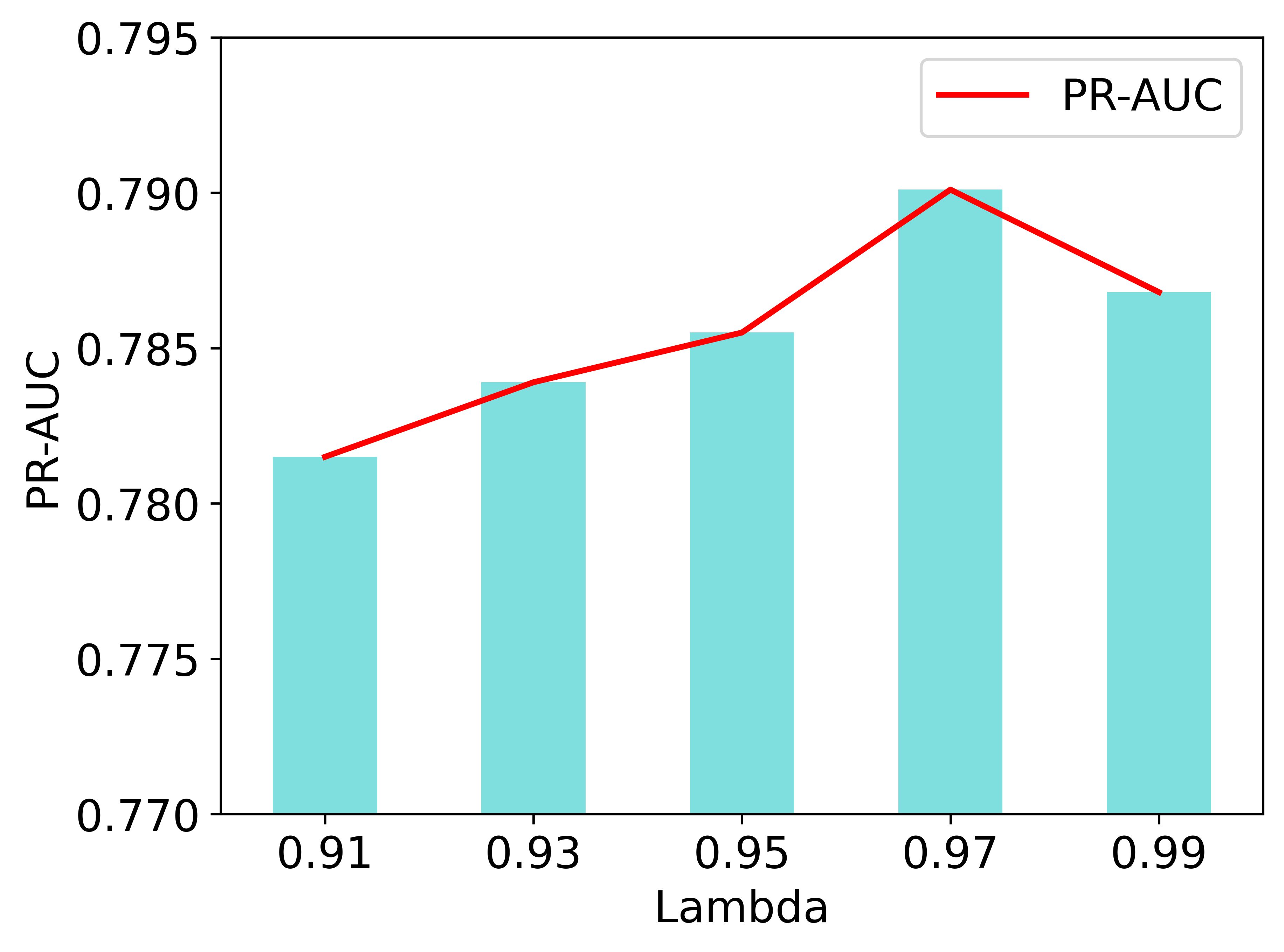}
	\end{subfigure}
	\hfill
	\begin{subfigure}[b]{0.3\linewidth}
		\centering
		\includegraphics[width=\linewidth]{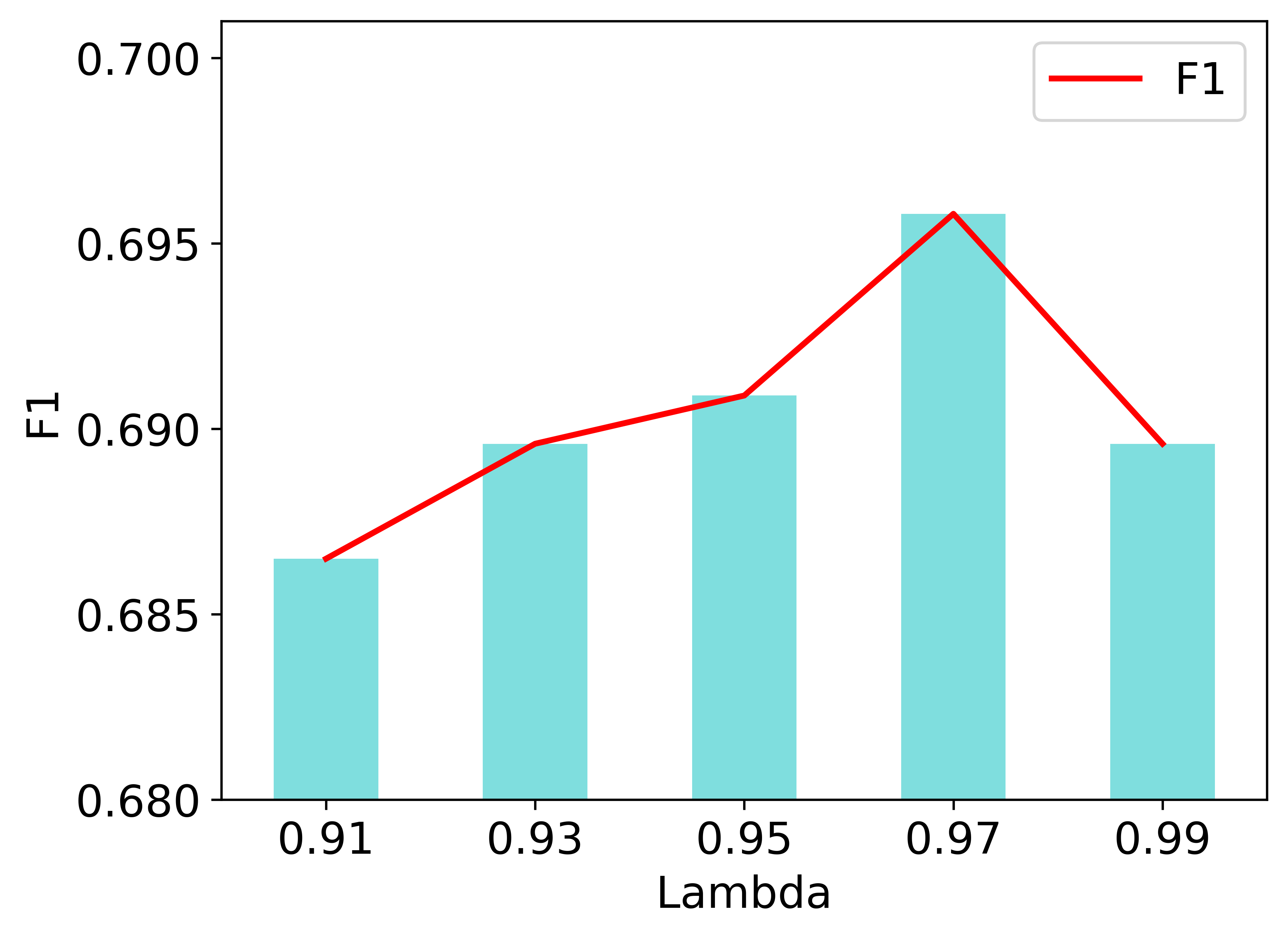}
	\end{subfigure}
	\caption{Impact of $\lambda$ values on MIMIC-III Dataset}
	\label{lambda}
\end{figure}

Firstly, with a fixed embedding dimension of 64, we explored different $\lambda$ values. As shown in Figure \ref{lambda}, the Jaccard, PR-AUC, and F1 scores reach their peak when $\lambda$ is set to 0.97.

\begin{figure}[h]
	\centering
	\begin{subfigure}[b]{0.3\linewidth}
		\centering
		\includegraphics[width=\linewidth]{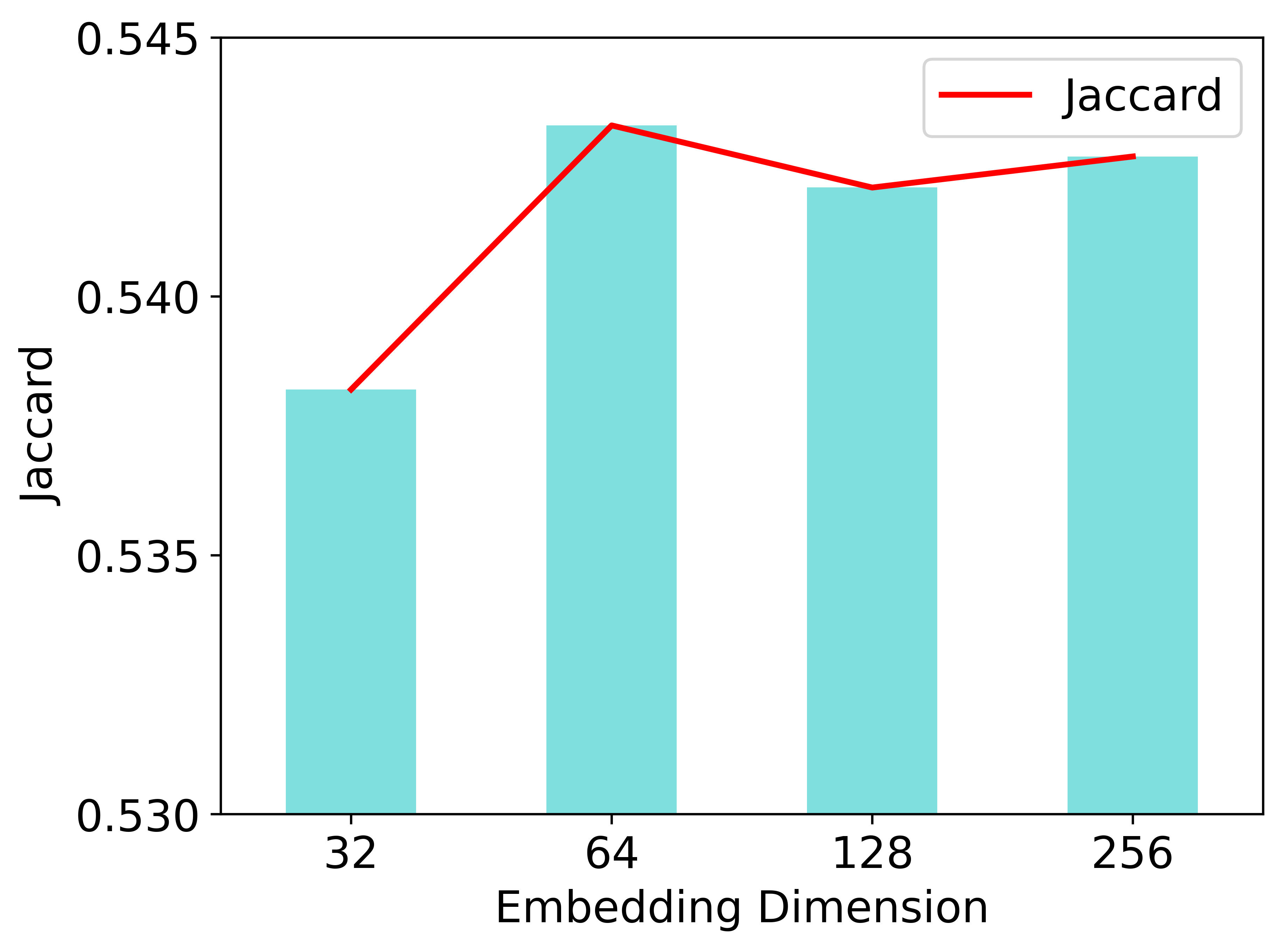}
	\end{subfigure}
	\hfill
	\begin{subfigure}[b]{0.3\linewidth}
		\centering
		\includegraphics[width=\linewidth]{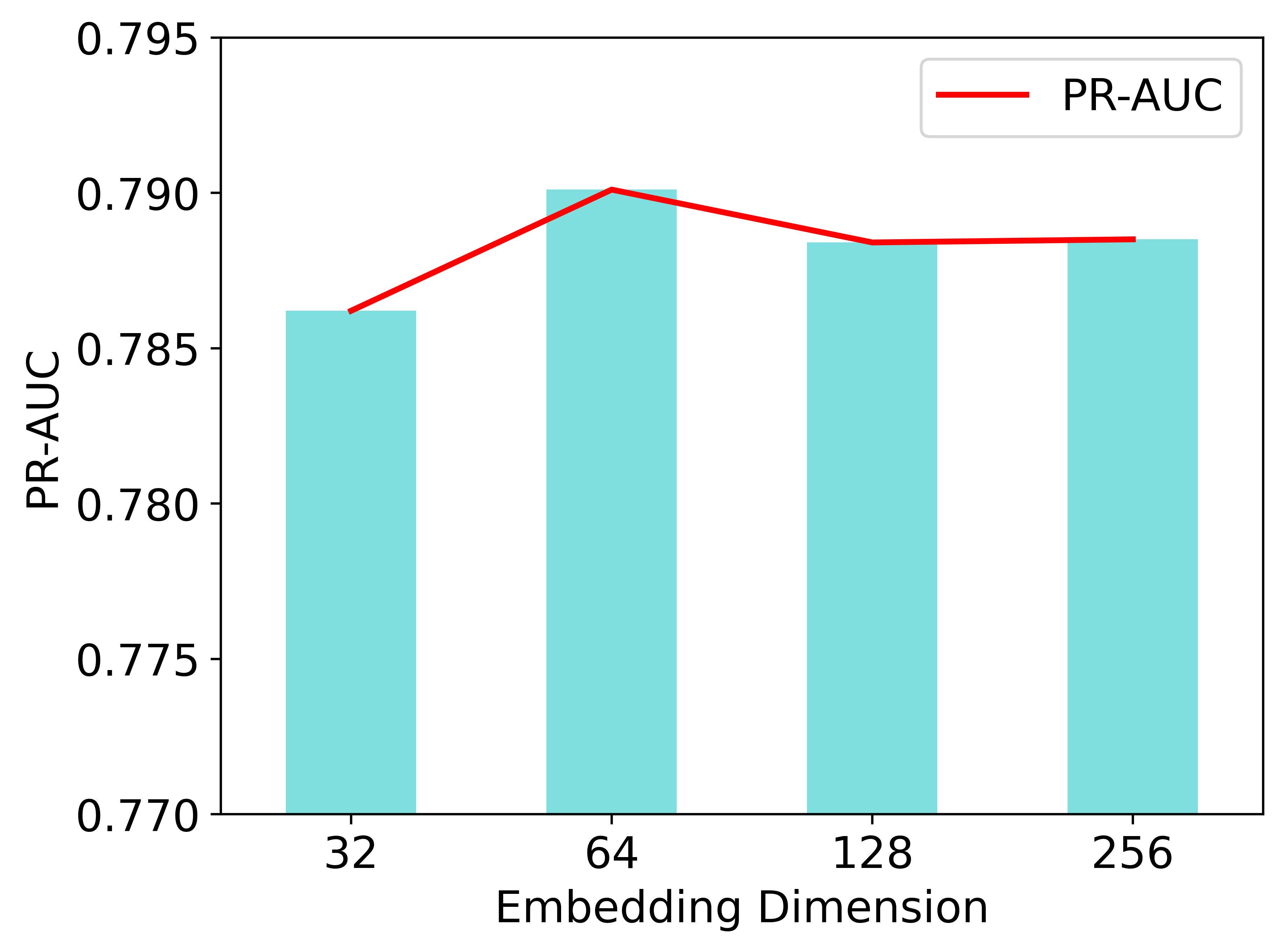}
	\end{subfigure}
	\hfill
	\begin{subfigure}[b]{0.3\linewidth}
		\centering
		\includegraphics[width=\linewidth]{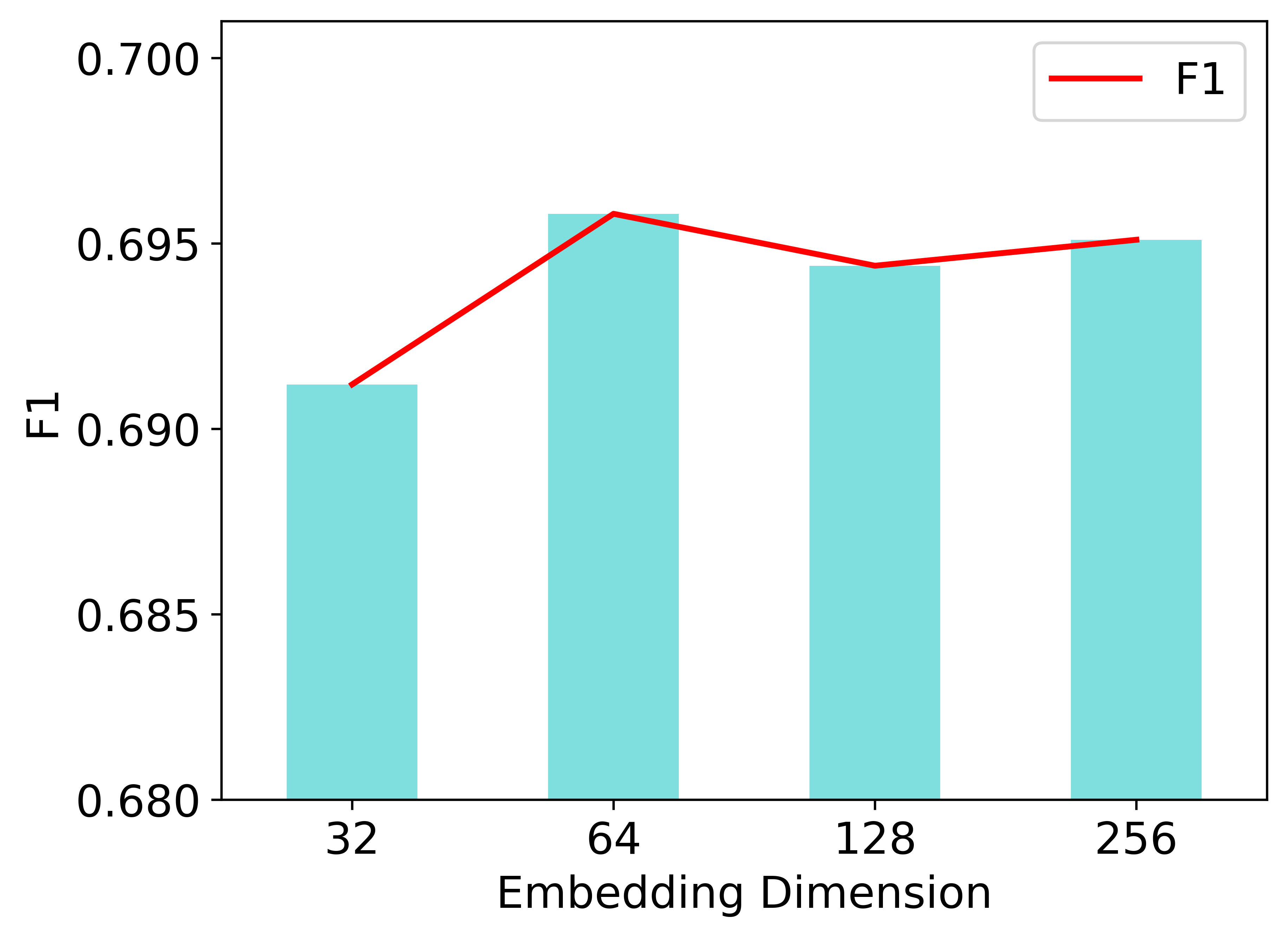}
	\end{subfigure}
	\caption{Impact of Embedding Dimensions on MIMIC-III Dataset}
	\label{dim}
\end{figure}

Subsequently, with a fixed learning $\lambda$ value of 0.97, we examined the impact of different embedding dimensions. The results, depicted in Figure \ref{dim}, indicate that an embedding dimension of 64 yields the highest Jaccard, PR-AUC, and F1 scores.

\subsection{Ablation study}
In the section, we removed some components of the ACDNet model to observe their impact on model performance.
\begin{itemize}
	\item[$\bullet$] ACDNet ${w/o \ M_e,M_d}$: We remove the EHR graph and DDI graph from input.
	\item[$\bullet$] ACDNet ${w/o \ M_{mol}}$: We remove the medicine molecular from input.
	\item[$\bullet$] ACDNet ${w/o \ att}$: We change the all self-attention to mean operation.
	\item[$\bullet$] ACDNet ${w/o \ seq^{att}_e}$ We remove the patient representations that have undergone self-attention.
	\item[$\bullet$] ACDNet ${w/o \ seq^{tr}_e}$ We remove the patient representations that have undergone Transformer Enconder.
	\item[$\bullet$] ACDNet ${w \ GRU}$: We replace the Transformer encoder with GRU.
	\item[$\bullet$] ACDNet ${w \ RNN}$: We replace the Transformer encoder with RNN.
	\item[$\bullet$] ACDNet ${w \ o_{1}}$: We remove vectors related to the medication record, such as $seq_m^{tr}$, $r_m$, $M$, $s_m$, and $o_{2}$, retaining only the patient EHR representation for recommendation.
\end{itemize}
\begin{table}[h]
	\centering
	\caption{Ablation Study on MIMIC-III}\label{tab4}
	\renewcommand{\arraystretch}{1.2}
	\setlength{\tabcolsep}{2pt}
	\scriptsize
	\begin{tabular}{lccccc}
		\hline
		Models & Jaccard & PR-AUC & F1     & DDI    & AVG\_Med \\ \hline
		ACDNet ${w/o \ M_e,M_d}$  & 0.5423 ± 0.0015 & 0.7888 ± 0.0022 & 0.6947 ± 0.0022 & 0.0861 ± 0.0009 & 20.4650 ± 0.1077   \\
		ACDNet ${w/o \ M_{mol}}$ & 0.5398 ± 0.0022 & 0.7861 ± 0.0024 & 0.6927 ± 0.0021 & 0.0817 ± 0.0008 & 20.8585 ± 0.1200   \\
		ACDNet ${w/o \ att}$ & 0.5350 ± 0.0016 & 0.7845 ± 0.0020 & 0.6887 ± 0.0016 & 0.0815 ± 0.0009 & 20.7952 ± 0.1420   \\
		ACDNet ${w/o \ seq^{att}_e}$ & 0.5001 ± 0.0029  & 0.7430 ± 0.0029  & 0.6568 ± 0.0026  & 0.0914 ± 0.0007  & 22.0030 ± 0.1511    \\
		ACDNet ${w/o \ seq^{tr}_e}$ & 0.5357 ± 0.0020   & 0.7834 ± 0.0014 & 0.6893 ± 0.0019 & \textbf{0.0814 ± 0.0007}  & 21.2701 ± 0.1517  \\
		ACDNet ${w \ GRU}$ & 0.5355 ± 0.0022 & 0.7814 ± 0.0023 & 0.6888 ± 0.0020 & 0.0861 ± 0.0009 & 21.1938 ± 0.1440   \\
		ACDNet ${w \ RNN}$ & 0.5323 ± 0.0024 & 0.7762 ± 0.0025 & 0.6860 ± 0.0022 & 0.0844 ± 0.0010 & 21.2428 ± 0.1564   \\
		ACDNet ${w \ o_{1}}$ & 0.5297 ± 0.0025 & 0.7789 ± 0.0024 & 0.6840 ± 0.0023 & 0.0889 ± 0.0008 & 20.0775 ± 0.0993   \\ 
		ACDNet  & \textbf{0.5433 ± 0.0027} & \textbf{0.7904 ± 0.0021} & \textbf{0.6957 ± 0.0021} & 0.0859 ± 0.0010 & 20.4900 ± 0.1197  \\ \hline
	\end{tabular}
\end{table}

The results from Table 6 indicate that the removal of the EHR graph and DDI graph has an observable impact on ACDNet's performance, suggesting the important role these two graphs may play in medication representation. Furthermore, the inclusion of medication molecular structure information also contributes to the effectiveness of medication representation. When the self-attention mechanism is eliminated from the ACDNet model, a noticeable decline in performance is observed, underscoring the critical role of the attention mechanism in the model.  After removing patient representations processed by self-attention, the model's performance significantly deteriorated. This indicates the vital role of self-attention in capturing each visit in a patient's history. The results obtained by removing the Transformer or replacing it with RNN or GRU in ACDNet emphasize the superiority of the Transformer encoder in comprehensively capturing essential information from patients' visit records and medication records. Moreover, when the collaborative decision-making component is removed, leaving only the direct recommendation part, a significant reduction in performance is evident. This emphasizes the significance of similarity-based collaborative decision-making in the model's overall effectiveness.

Overall, these findings demonstrate the importance of incorporating EHR and DDI graphs, attention mechanism, Transformer encoder, and collaborative decision-making in ACDNet to achieve superior performance in medication recommendation tasks.

\subsection{Case study}

To investigate the effectiveness of ACDNet under different numbers of visits, we selected two patients, one with 2 visit records and the other with 5 visit records. We employed four models, namely SafeDrug, COGNet, FFBDNet, and ACDNet, for medication recommendations. Figures 4 and 5 display the recommendation results, where "Correct" refers to medications that were accurately recommended. "Unseen" denotes medications predicted by the model but not found in the actual data, while "Missed" indicates medications present in the actual data that the model failed to predict.

For patient A, who had only two visit records, ACDNet displayed remarkable precision in medication recommendations, and the occurrence of unseen medications was notably minimal. This underscores the robust effectiveness of ACDNet in medication recommendations, as it consistently minimizes the occurrence of erroneous medication suggestions.
\begin{figure}[h]
	\centering
	\begin{subfigure}[b]{0.3\linewidth}
		\centering
		\includegraphics[width=\linewidth]{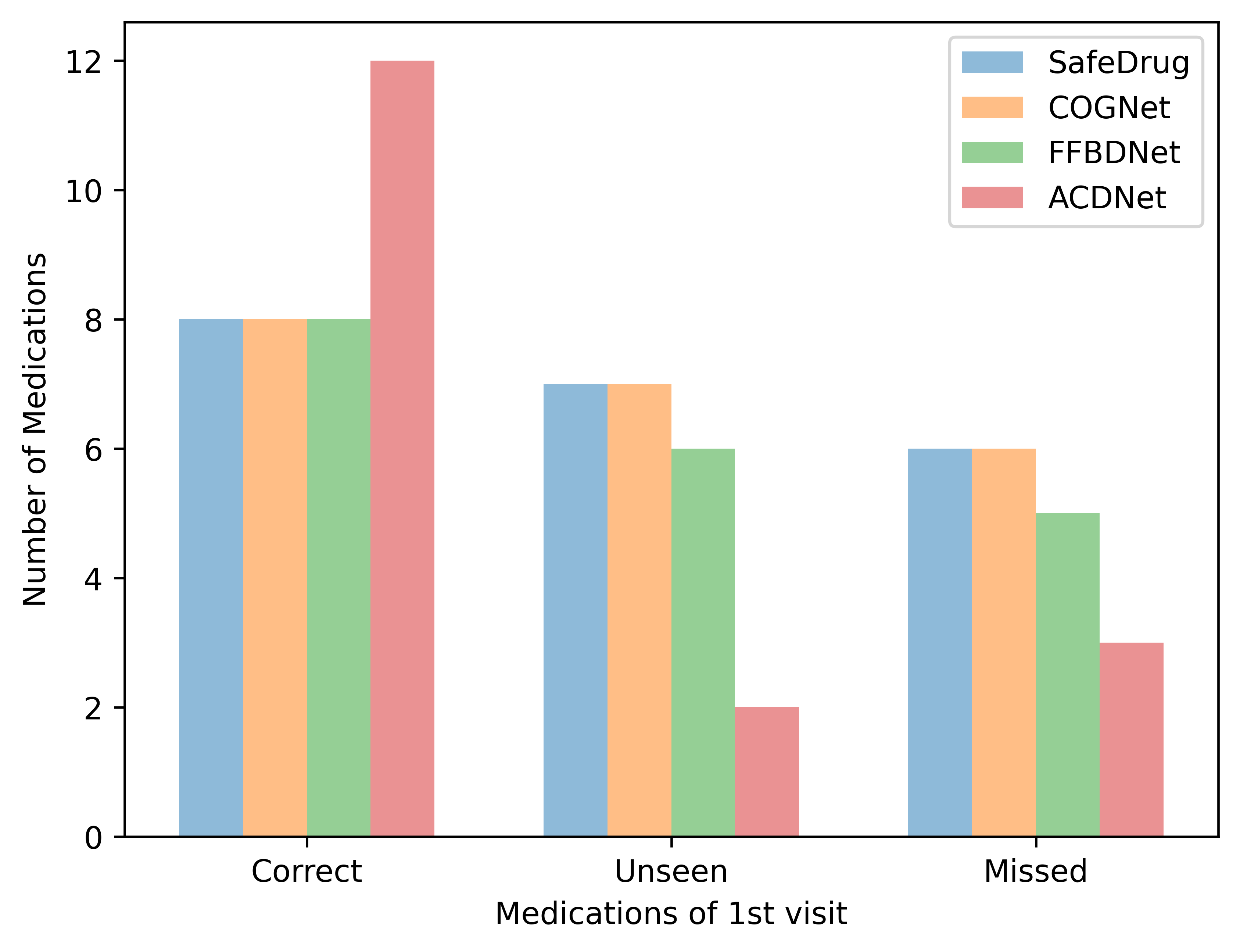}
	\end{subfigure}
	\hspace{1cm}
	\begin{subfigure}[b]{0.3\linewidth}
		\centering
		\includegraphics[width=\linewidth]{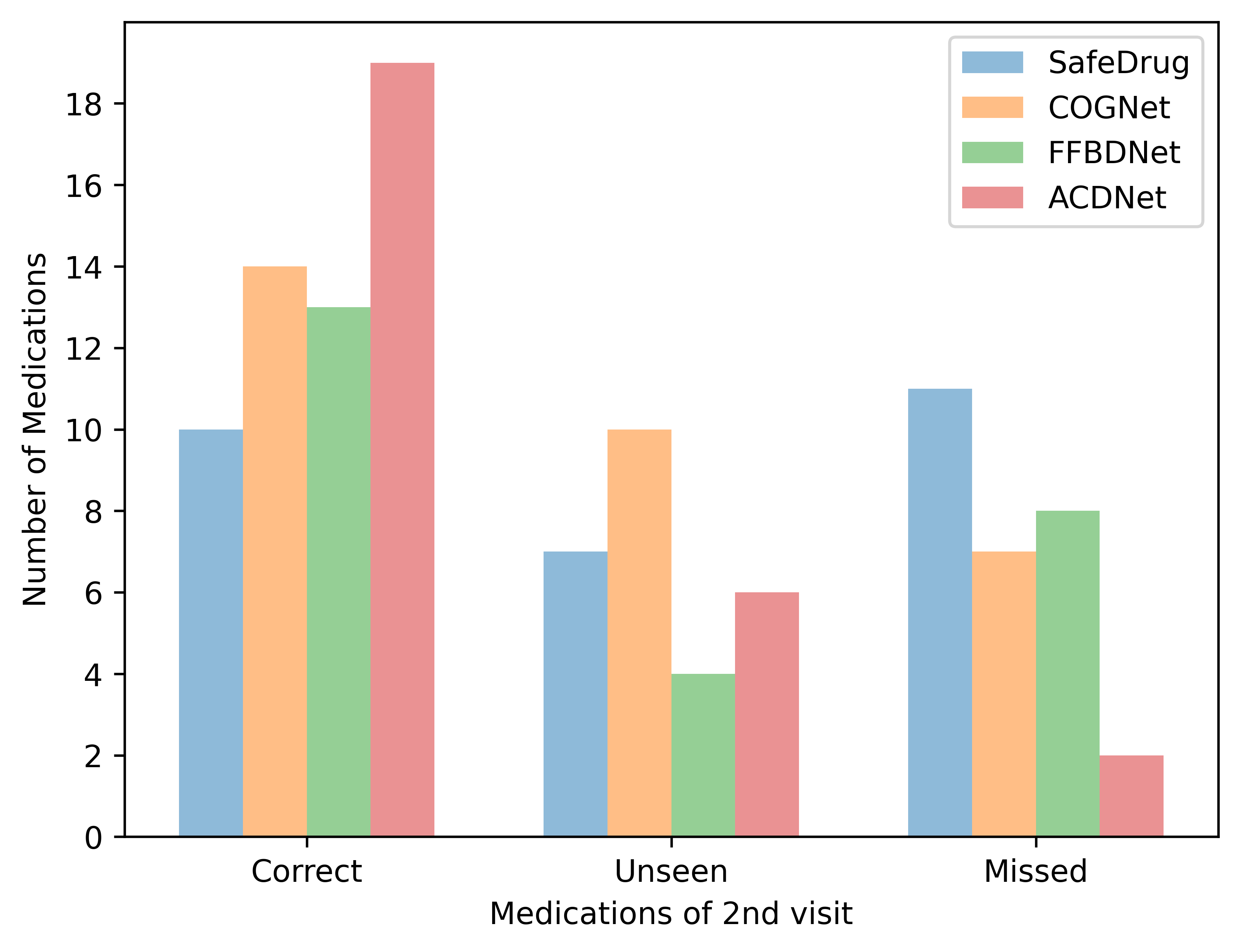}
		
	\end{subfigure}
	\caption{Recommended Medications of Patient A}
	\label{patientA}
\end{figure}

In the scenario of patient B, who had a more extensive medical history with five visit records, ACDNet maintained its exceptional performance. It delivered a considerably higher number of accurate medication recommendations when compared to other models. This emphasizes the model's consistent excellence in catering to patients with a richer medical journey, establishing ACDNet as a reliable choice for medication recommendation across diverse patient profiles.
\begin{figure}[h]
	\centering
	\begin{subfigure}[b]{0.3\linewidth}
		\centering
		\includegraphics[width=\linewidth]{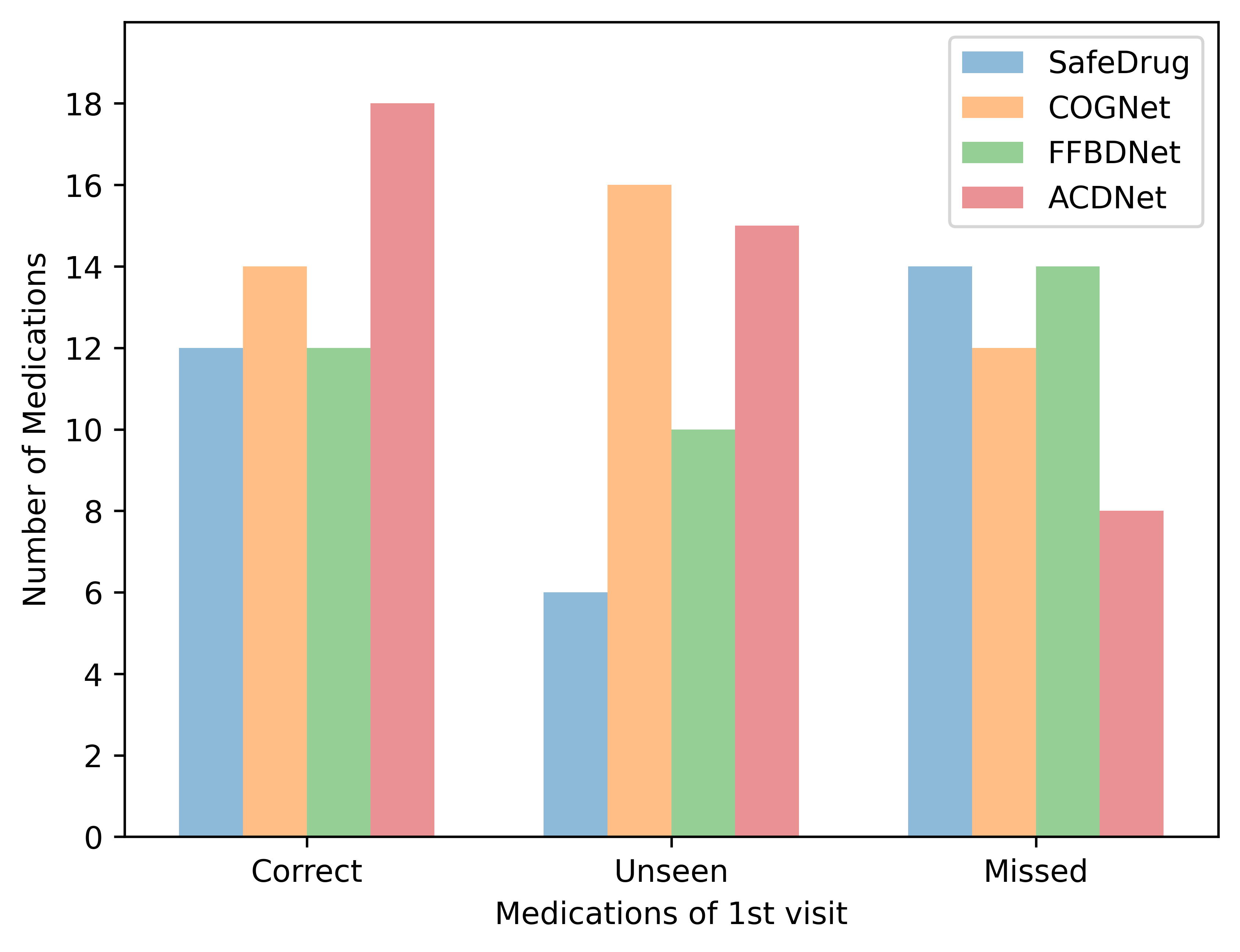}
	\end{subfigure}
	\hfill
	\begin{subfigure}[b]{0.3\linewidth}
		\centering
		\includegraphics[width=\linewidth]{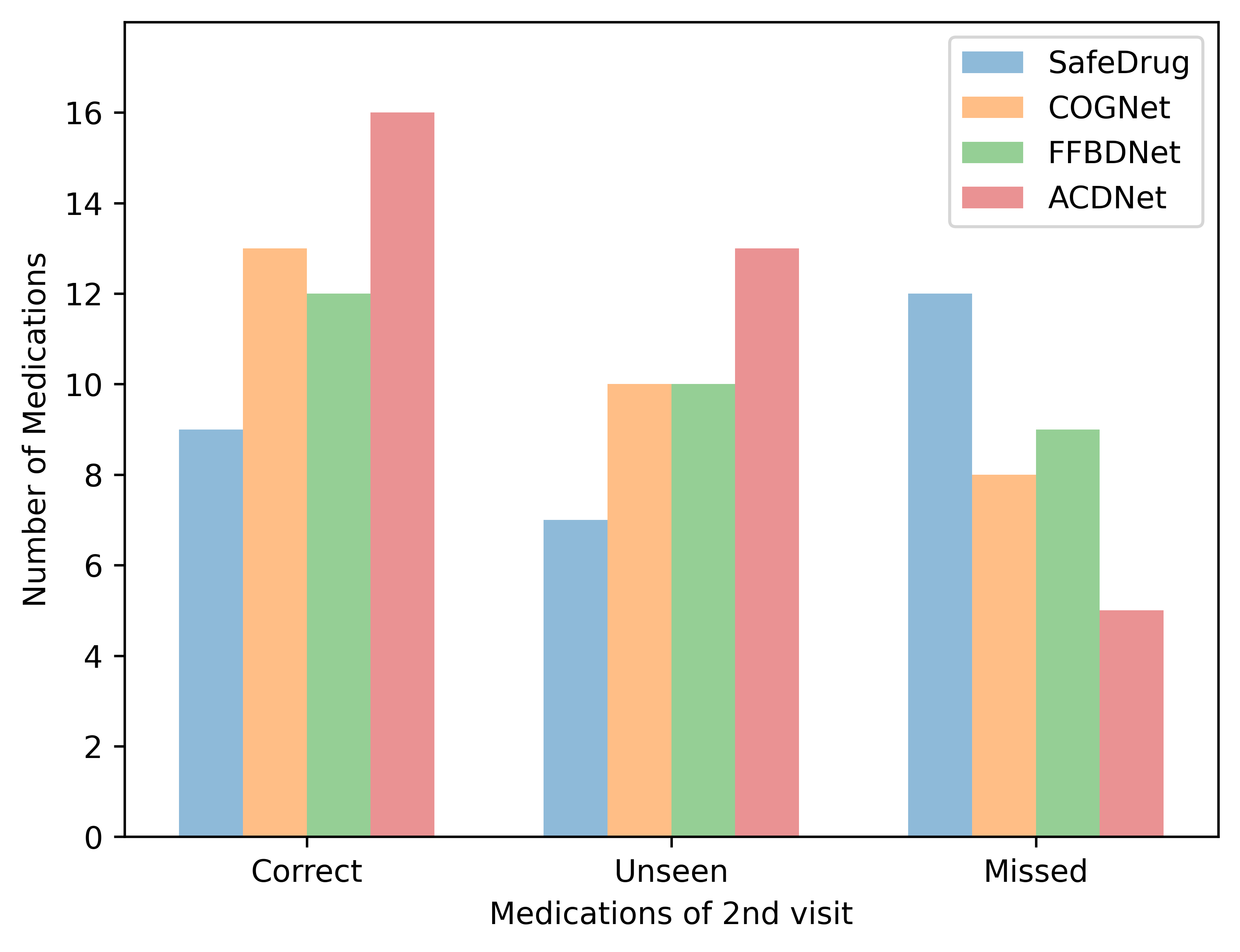}
	\end{subfigure}
	\hfill
	\begin{subfigure}[b]{0.3\linewidth}
		\centering
		\includegraphics[width=\linewidth]{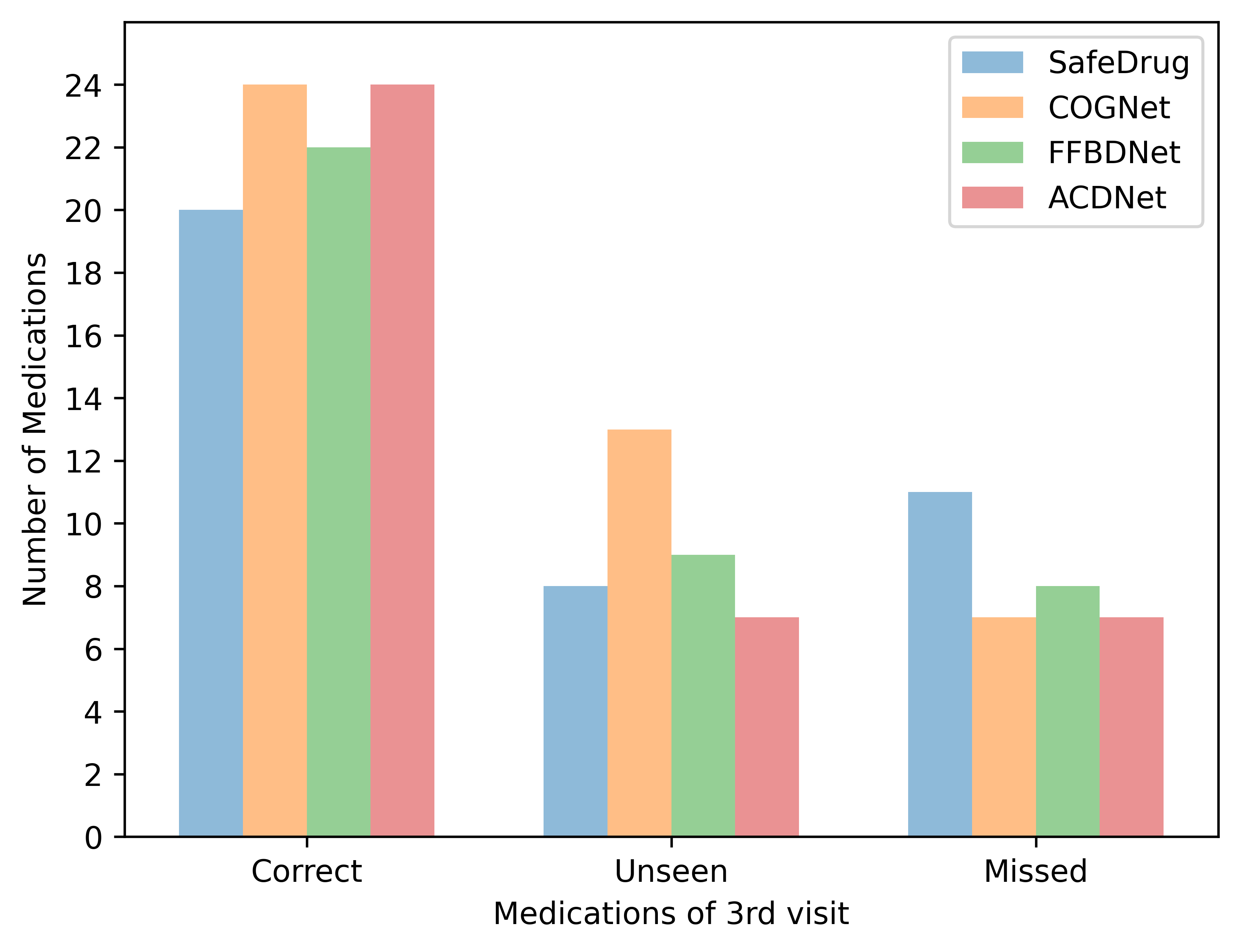}
	\end{subfigure}
	\hfill
	\begin{subfigure}[b]{0.3\linewidth}
		\centering
		\includegraphics[width=\linewidth]{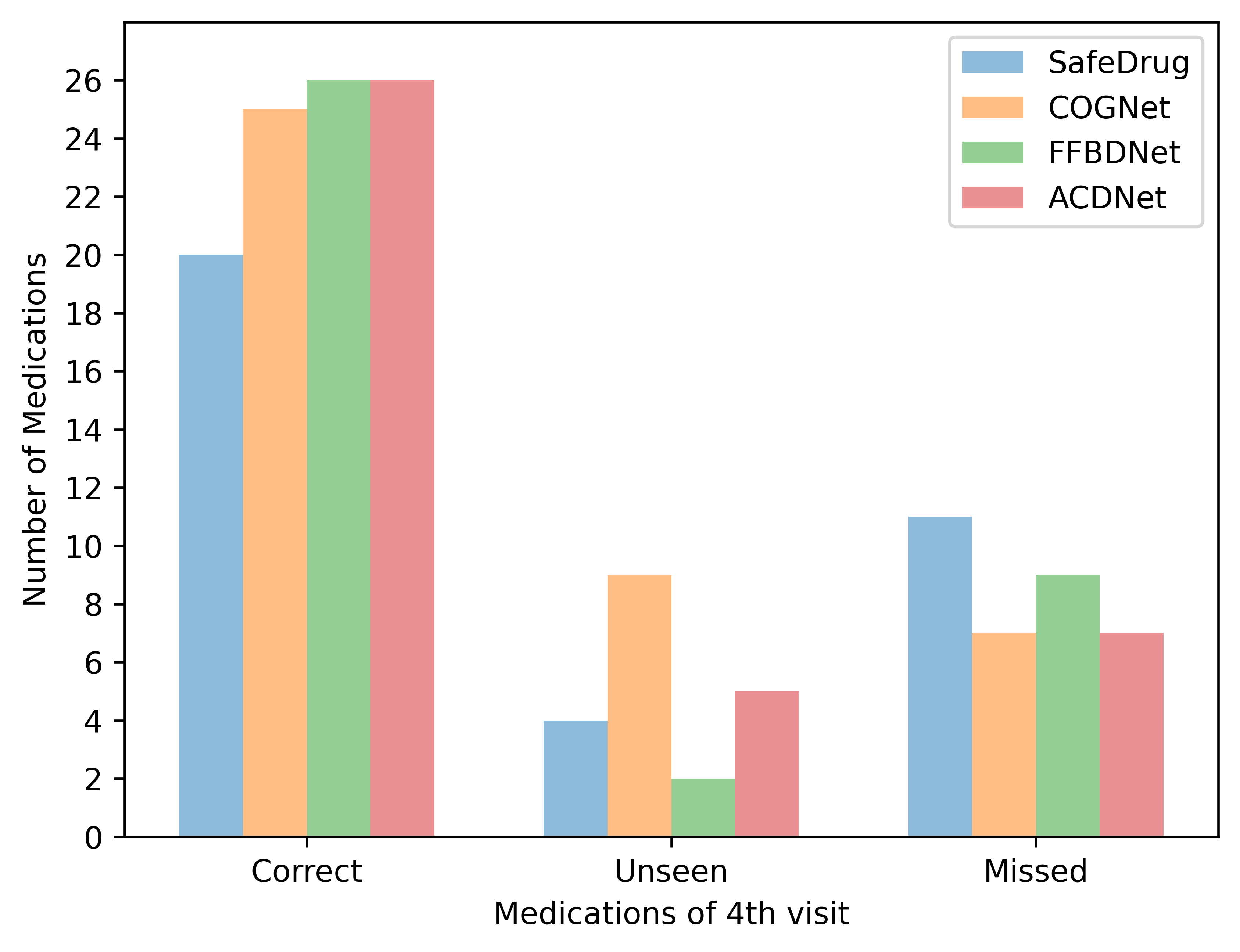}
	\end{subfigure}
	\hspace{1cm}
	\begin{subfigure}[b]{0.3\linewidth}
		\centering
		\includegraphics[width=\linewidth]{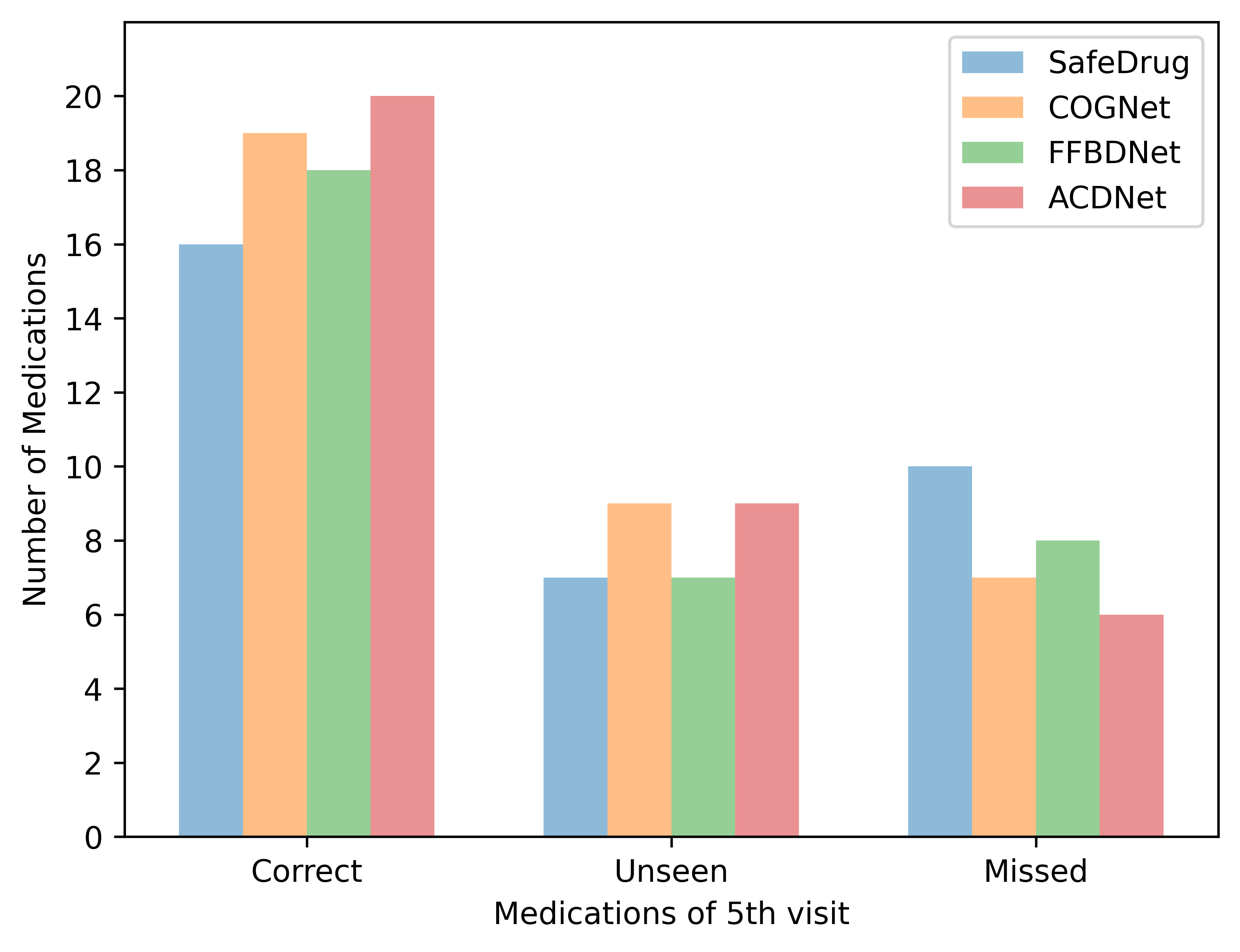}
	\end{subfigure}
	
	\caption{Recommended Medications of Patient B}
	\label{patientB}
\end{figure}
\section{Conclusion}
Medication recommendation constitutes a crucial application of data mining within the biomedical domain. However, existing models grapple with inadequacies concerning patient information and medicine representation. To address these issues, this study introduces the ACDNet model for modeling and predicting patients' health and medicine information. The model employs attention mechanism and Transformer to encode patients' EHR data and incorporates various external knowledge sources to represent medicine information. By combining the collaborative decision mechanism based on medication record similarity, we have successfully achieved personalized medication recommendation. We experimented with MIMIC-III and MIMIC-IV datasets, comparing ACDNet to other models. Results show ACDNet outperforms in Jaccard, PR-AUC, and F1 metrics, proving its efficacy in acquiring patient health data and medication representations. In future work, we will strive to reduce the interaction rate between drugs to avert potential adverse reactions and side effects. 

\section*{CRediT authorship contribution statement}
\textbf{Jiacong Mi:} Methodology, Software, Validation, Writing - original draft. \textbf{Yi Zu:} Data curation,  Writing - review \& editing. \textbf{Zhuoyuan Wang:} Data curation,  Writing - review \& editing.  \textbf{Jieyue He:} Supervision,  Writing - review \& editing.
\section*{Declaration of Competing Interest}
The authors declare that they have no known competing financial interests or personal relationships that could have appeared to influence the work reported in this paper.
\section*{Acknowledgments}
This work was supported by National Key R\&D Program of China (2019YFC1711000) and Collaborative Innovation Center of Novel Software Technology and Industrialization. 

\bibliographystyle{elsarticle-num} 
\biboptions{square,numbers,sort&compress}
\bibliography{ACDNet.bib}

\end{document}